\documentclass[10pt,journal,compsoc]{IEEEtran}

\ifCLASSOPTIONcompsoc
  \usepackage[nocompress]{cite}
\else
  \usepackage{cite}
\fi

\ifCLASSINFOpdf
\else
\fi

\usepackage{marvosym,hyperref,amsmath,amssymb,xcolor,ragged2e}
\usepackage{booktabs,multirow,multicol,caption,makecell,graphicx,float}

\begin{document}

\title{MMRel: Benchmarking Relation Understanding in Multi-Modal Large Language Models}

\author{Jiahao Nie$^*$, \ 
        Gongjie Zhang$^*$, \ 
        Wenbin An$^*$, \ 
        Yun Xing$^*$,\\
        Yap-Peng Tan,~\IEEEmembership{Fellow,} \ 
        Alex C. Kot,~\IEEEmembership{Life Fellow,} \ 
        and \ Shijian Lu\textsuperscript{\Letter}
\IEEEcompsocitemizethanks{
\IEEEcompsocthanksitem Jiahao Nie is with the Interdisciplinary Graduate Programme, Nanyang Technological University, Singapore, and the Nanyang Technological University, Singapore.
\IEEEcompsocthanksitem Gongjie Zhang is with the Alibaba DAMO Academy, Singapore.
\IEEEcompsocthanksitem Wenbin An is with the Xi'an Jiaotong University, China.
\IEEEcompsocthanksitem Yun Xing, Alex C. Kot, and Shijian Lu are with the Nanyang Technological University, Singapore.
\IEEEcompsocthanksitem Yap-Peng Tan is with the VinUniversity, Vietnam, and the Nanyang Technological University, Singapore.
\IEEEcompsocthanksitem Email: jiahao007@e.ntu.edu.sg \ \ \ shijian.lu@ntu.edu.sg
\IEEEcompsocthanksitem *: Equal contribution. \ \ \ \Letter: Corresponding author.}
}

\markboth{Journal of \LaTeX\ Class Files}%,~Vol.~14, No.~8, August~2015}%
{Shell \MakeLowercase{\textit{et al.}}: Bare Demo of IEEEtran.cls for Computer Society Journals}

\IEEEtitleabstractindextext{%
\begin{abstract}
\justifying
% Though Multi-modal Large Language Models (MLLMs) have recently achieved significant progress, they often face various problems while handling inter-object relations. Nevertheless, the lack of sufficient training and evaluation data for relation understanding has greatly hindered the progress of MLLMs in various vision–language perception tasks. 
Though Multi-modal Large Language Models (MLLMs) have recently achieved significant progress, they often struggle to understand diverse and complicated inter-object relations. Specifically, the lack of large-scale and high-quality relation data has greatly hindered the progress of MLLMs in various vision–language perception tasks. We attempt to address this challenge by contributing \textbf{M}ulti-\textbf{M}odal \textbf{Rel}ation Understanding benchmark (\textbf{MMRel}) that features large-scale, high-quality, and diverse data on inter-object relations. MMRel features three distinctive attributes: (i) It contains 22{,}500 question-answer pairs, spanning three distinct domains and around 400 relations, ensuring both scale and diversity; (ii) it provides manually verified, high-quality labels to ensure exceptional annotation accuracy; (iii) it includes adversarial cases with highly unusual relations, offering a challenging setting for evaluating relation hallucination. These features make MMRel ideal for evaluating MLLMs on relation understanding, as well as for fine-tuning MLLMs to enhance relation comprehension capability. Extensive experiments on \textbf{28} MLLMs demonstrate the effectiveness of MMRel in both evaluating and enhancing MLLMs’ relation understanding, and the accompanying analyses provide insights for future research. The benchmark has been made publicly available \href{https://niejiahao1998.github.io/MMRel}{\textcolor{magenta}{here}}.
\end{abstract}
}

\maketitle

\IEEEdisplaynontitleabstractindextext

\IEEEpeerreviewmaketitle

\IEEEraisesectionheading{\section{Introduction}\label{sec:intro}}

\begin{figure}[t]
\centering
    % \vspace{-1mm}
    \includegraphics[width=\linewidth]{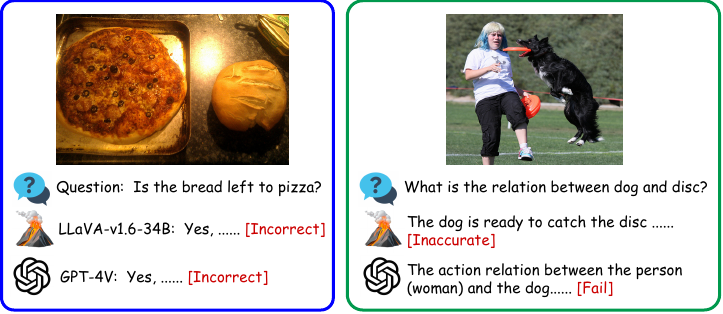}
    \vspace{-4mm}
    \caption{Multi-Modal Large Language Models tend to fail in understanding inter-object relations.}
    % \vspace{-1mm}
    \label{fig:fail}
\end{figure}

\IEEEPARstart{M}{ulti-Modal} Large Language Models (MLLMs)~\cite{liu2023visual,liu2023improved,ye2023mplugowl,li2023otter,dai2023instructblip,gong2023multimodalgpt,bai2023qwen,yu2024rlhf,you2023ferret,guo2024regiongpt} have demonstrated remarkable capabilities in various vision-language tasks~\cite{lu2022learn,lu2023mathvista,yue2024mmmu,yu2023mm,li2024seed,liu2023mmbench,chen2024we,an2024agla,xing2024mitigating,chen2024multi,chen2024alleviating,wang2024all,qiu2024longhalqa}. One essential ability for these tasks is understanding the complex relations among objects in images~\cite{krishna2017visual,zhao2022vl}, which is crucial for advanced VQA~\cite{lee2024visual,wang2024weakly,an2024knowledge} and detailed image captioning~\cite{li2023blip,wang2023caption}. However, recent studies~\cite{zhao2022vl,doveh2023teaching,liu2024survey,bai2024hallucination} reveal that most existing MLLMs tend to struggle in interpreting inter-object relations, as illustrated in Fig.~\ref{fig:fail}. This challenge highlights the need for a large-scale, diverse, and high-quality benchmark tailored for MLLMs, representing a crucial step toward assessing their limitations and enhancing their relation understanding.

\begin{figure}[t]
\centering
    \vspace{-3mm}
    \includegraphics[width=\linewidth]{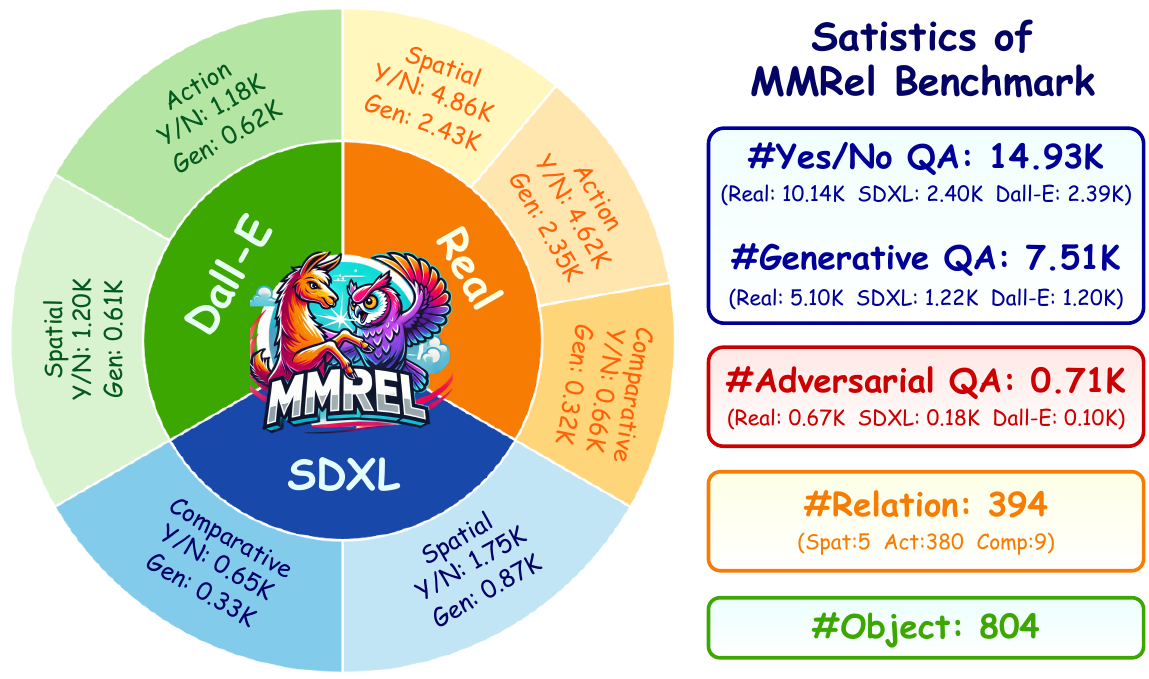}
    \vspace{-3mm}
    \caption{The proposed Multi-Modal Relation Understanding benchmark (MMRel) features large-scale, high-quality, and diverse data. \textbf{Left}: MMRel covers three domains that span seven subsets. \textbf{Right}: MMRel comprises 22{,}500 QA pairs on 804 objects and 394 kinds of relations.}
    % \vspace{+1mm}
    \label{fig:statistics}
\end{figure}

\begin{table*}[t]
    
\captionsetup{justification=justified,singlelinecheck=false}
\caption{A comparison of MMRel with existing benchmarks on inter-object relations. MMRel is featured with large-scale, diverse, and high-quality data on inter-object relations. Specifically, MMRel spans three distinct domains and three categories of inter-object relations including an adversarial subset. Its 22{,}500 QAs comprise both \textbf{discriminative} \textit{Yes/No} and \textbf{generative} \textit{open-ended} questions.}
\vspace{-1mm}

\centering
\renewcommand\arraystretch{1.25}

\resizebox{\linewidth}{!}{
    \begin{tabular}{l||c|ccc|ccc|c|c|c|c}
        \toprule
        \multirow{2}*{\textbf{Benchmark}} & \multirow{2}*{\textbf{\#Q\&A}} & \multicolumn{3}{c|}{\textbf{Taxonomy}} & \multicolumn{3}{c|}{\textbf{Domain}} & \textbf{Human} & \textbf{Adv.} & \textbf{Dis.} & \textbf{Gen.} \\ \cline{3-8}
        ~ & ~ & \textbf{\textit{Spatial}} & \textbf{\textit{Action}} & \textbf{\textit{Comparative}} & \textbf{\textit{Real}} & \textbf{\textit{SDXL}} & \textbf{\textit{Dall-E}} & \textbf{Check} & \textbf{Subset} & \textbf{QA} & \textbf{QA} \\ \hline\hline
        MMHAL~\cite{sun2023aligning} & 12 & $\checkmark$ & $\times$ & $\times$ & $\checkmark$ & $\times$ & $\times$ & $\checkmark$ & $\times$ & $\times$ & $\checkmark$ \\
        RAH~\cite{chen2023mitigating} & 500 & $\checkmark$ & $\checkmark$ & $\times$ & $\checkmark$ & $\times$ & $\times$ & $\times$ & $\times$ & $\checkmark$ & $\times$  \\
        MERLIM~\cite{villa2023behind} & - & $\checkmark$ & $\times$ & $\times$ & $\checkmark$ & $\times$ & $\times$ & $\times$ & $\times$ & $\checkmark$ & $\times$ \\
        FAITHSCORE~\cite{jing2023faithscore} & - & - & - & - & $\checkmark$ & $\times$ & $\times$ & $\checkmark$ & $\times$ & $\times$ & $\checkmark$ \\
        M-HalDetect~\cite{gunjal2024detecting} & - & - & - & - & $\checkmark$ & $\times$ & $\times$ & $\checkmark$ & $\times$ & $\times$ & $\checkmark$ \\
        AMBER~\cite{wang2023llm} & 1.7K & $\checkmark$ & $\times$ & $\times$ & $\checkmark$ & $\times$ & $\times$ & $\checkmark$ & $\times$ & $\checkmark$ & $\times$ \\
        SPEC~\cite{peng2023synthesize} & 3.5K & $\checkmark$ & $\times$ & $\checkmark$ & $\times$ & $\checkmark$ & $\times$ & $\times$ & $\times$ & $\checkmark$ & $\times$ \\
        Hallusion~\cite{liu2023hallusionbench} & 150 & $\times$ & $\times$ & $\checkmark$ & $\checkmark$ & $\times$ & $\times$ & $\times$ & $\times$ & $\checkmark$ & $\times$ \\
        MME~\cite{peng2023synthesize} & 60 & $\checkmark$ & $\times$ & $\times$ & $\checkmark$ & $\times$ & $\times$ & $\times$ & $\times$ & $\checkmark$ & $\times$ \\
        R-Bench~\cite{wu2024evaluating} & 11.7K & $\checkmark$ & $\checkmark$ & $\times$ & $\checkmark$ & $\times$ & $\times$ & $\checkmark$ & $\times$ & $\checkmark$ & $\times$ \\
        \textbf{MMRel} & \textbf{22.5K} & \textbf{$\checkmark$} & \textbf{$\checkmark$} & \textbf{$\checkmark$} & \textbf{$\checkmark$} & \textbf{$\checkmark$} & \textbf{$\checkmark$} & \textbf{$\checkmark$} & \textbf{$\checkmark$} & \textbf{$\checkmark$} & \textbf{$\checkmark$} \\ \bottomrule
    \end{tabular}
}

\vspace{-2mm}
\label{tab:exitsing_dataset}
\end{table*}

Though several benchmarks involving inter-object relations have been proposed, none of them were designed specifically to assess the relation understanding capabilities of MLLMs~\cite{wang2023llm, sun2023aligning, peng2023synthesize,gunjal2024detecting, chen2023mitigating,villa2023behind, jing2023faithscore, fu2023mme,liu2023hallusionbench,wu2024evaluating}. As summarized in Tab.~\ref{tab:exitsing_dataset}, most existing benchmarks suffer from clear limitations in data scale, relation categories, and data diversity. To address these issues, we construct a comprehensive benchmark dedicated to inter-object relations, aiming to both evaluate and enhance the relation understanding capabilities of MLLMs. Specifically, we first taxonomize relations by three distinct categories, including spatial relations (\textit{e.g.}, a dog is \textbf{\textit{left}} to a cat), action relations (\textit{e.g.}, a boy \textbf{\textit{eats}} a burger), and comparative relations (\textit{e.g.}, an apple is \textbf{\textit{smaller}} than a watermelon). Moreover, we introduce \textit{relation hallucination}, a new concept referring to the scenario when MLLM-generated relations align with the common sense but contradict the actual contents in images (e.g., a MLLM responds by ``a man \textbf{\textit{drives}} a car'' but the actual image content is ``a man \textbf{\textit{pushes}} a car''.

\begin{figure*}[t]
\centering
   \vspace{-2mm}
   \includegraphics[width=\linewidth]{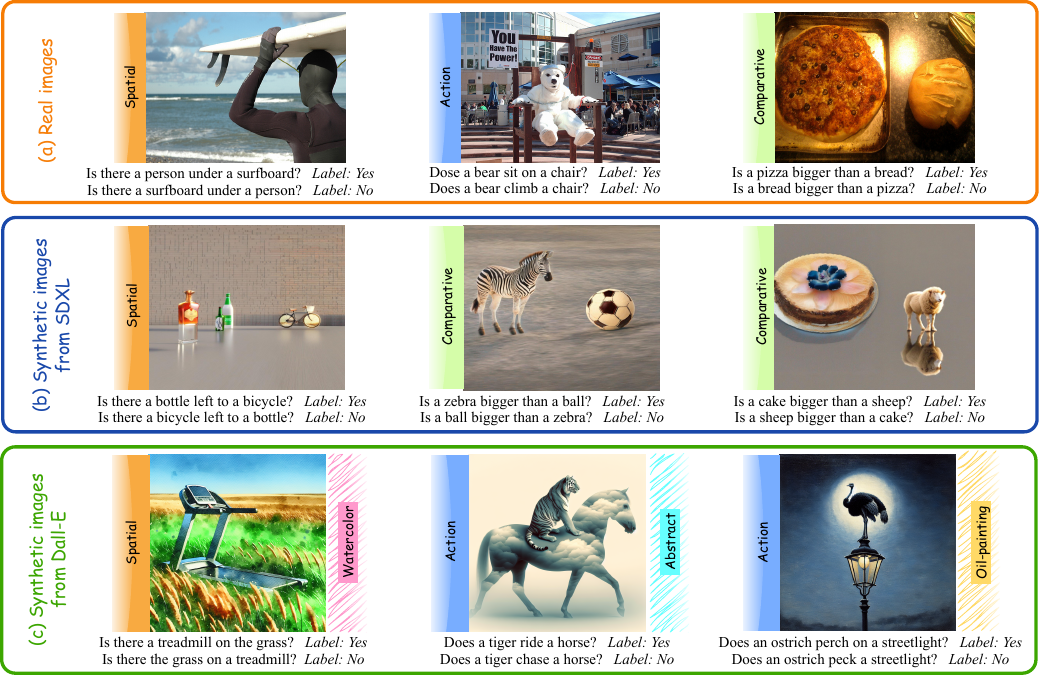}
   \vspace{-4mm}
   \caption{Sample images from the MMRel benchmark.
   MMRel consists of three categories of inter-object relations: spatial, action, and comparative relations. The images are sourced from three domains: (a) real images, (b) synthetic images generated by SDXL, and (c) images generated by Dall-E. More Dall-E samples are shown in Fig.~\ref{fig:dall-e_sample} of the appendix.}
   % \vspace{+3mm}
\label{fig:sample}
\end{figure*}

The constructed \textbf{M}ulti-\textbf{M}odal \textbf{Rel}ation Understanding benchmark (\textbf{MMRel}) has several desirable features, as illustrated in Fig.~\ref{fig:statistics}. \textbf{\textit{(i)}} It is large-scale, comprising multimodal data covering 394 types of inter-object relations across three categories (\textit{i.e.}, spatial, action, and comparative), along with a challenging adversarial subset containing abnormal relations that deviate from common sense. In total, the benchmark includes around 22{,}500 Q\&A pairs, each accompanied by clear annotations and relation labels. \textbf{\textit{(ii)}} It exhibits strong diversity, with image data sourced from three distinctive origins: real images, SDXL~\cite{podell2023sdxl}-generated images with clean backgrounds, and DALL\text{-}E~\cite{betker2023improving}-generated images in multiple styles, as illustrated in Fig.~\ref{fig:sample}. \textbf{\textit{(iii)}} It offers high-quality data collected through a semi-automatic pipeline. Specifically, we first generate images and draft annotations using MLLMs~\cite{betker2023improving,achiam2023gpt}, and subsequently verify and correct them through human review. \textit{\textbf{(iv)}} It contains both widely adopted \textbf{discriminative} questions with \textit{Yes/No} answers~\cite{wang2023llm, chen2023mitigating, villa2023behind,fu2023mme, liu2023hallusionbench,li2023evaluating,wu2024evaluating, zheng2024reefknot} and \textbf{generative} \textit{open-ended} questions through LLM-assisted evaluations~\cite{liu2023visual,sun2023aligning}.

MMRel can serve different purposes thanks to its large-scale, diverse, and high-quality features. \textit{\textbf{(i)}} Its primary use is to evaluate existing MLLMs in relation understanding. Extensive evaluations on 28 MLLMs reveal that most general MLLMs~\cite{liu2023improved,dai2023instructblip, bai2023qwen, yao2024minicpm,blog2024hello}, grounding-specific MLLMs~\cite{chen2023shikra,you2023ferret,cheng2024spatialrgpt}, and even hallucination mitigation approaches~\cite{chuang2023dola, huang2023opera, leng2023mitigating} still exhibit sub-optimal performance in understanding inter-object relations. \textit{\textbf{(ii)}} The evaluation results further provide a variety of insightful observations, revealing limitations of current models and suggesting promising directions for future research. \textit{\textbf{(iii)}} Another important application of MMRel is fine-tuning MLLMs to enhance their relation understanding capabilities~\cite{liu2023visual,liu2023improved}. Our experiments indicate that using MMRel as training data substantially improves both relation understanding and related perceptual capabilities.

\noindent Our contributions can be summarized in four aspects:\\
\noindent$\bullet$ \textit{\textbf{First}}, we systematically define inter-object relations and relation hallucinations under the context of MLLMs, laying a solid foundation for future study of relation hallucinations and their mitigation in MLLMs (Sec.~\ref{sec:intro}).\\
\noindent$\bullet$ \textit{\textbf{Second}}, we contribute MMRel, a large-scale benchmark with 22{,}500 high-quality vision-language samples sourced from diverse domains, each with clearly defined inter-object relations. To the best of our knowledge, this is the first MLLM benchmark dedicated to inter-object relations with superior diversity and quality (Sec.~\ref{sec:review} and Sec.~\ref{sec:mmrel}).\\
\noindent$\bullet$ \textit{\textbf{Third}}, we conduct extensive experiments on 28 MLLMs showing that MMRel is effective not only for evaluating existing MLLMs but also for fine-tuning them to enhance their relation understanding capabilities (Sec.~\ref{sec:exp}).\\ 
\noindent$\bullet$ \textit{\textbf{Fourth}}, we provide systematic analyses of the experimental results and offer key insights and takeaways for future research (Sec.~\ref{sec:exp} and Sec.~\ref{sec:takeaway}).
\section{Review of Prior Benchmarks for Relation Understanding}
\label{sec:review}

The study of inter-object relations boasts a rich history, with Scene Graph Generation (SGG)~\cite{johnson2015image,zellers2018neural,tang2020unbiased} serving as a foundational area of focus. SGG was collected for identifying objects and their relations within images. Leveraging specialized object detection methods~\cite{redmon2016you,lin2017focal} that excel in detecting both tiny and inconspicuous objects, SGG benchmarks such as Visual Genome benchmark (VG)~\cite{krishna2017visual} are distinguished by their extremely detailed and comprehensive annotations. With the recent successes of Multi-Modal Large Language Models (MLLMs)~\cite{radford2021learning}, several new benchmarks~\cite{doveh2023teaching, yuksekgonul2022and} were collected for evaluating the relation understanding capabilities of MLLMs.

\begin{figure*}[t]
\centering
    \vspace{-2mm}
    \includegraphics[width=\linewidth]{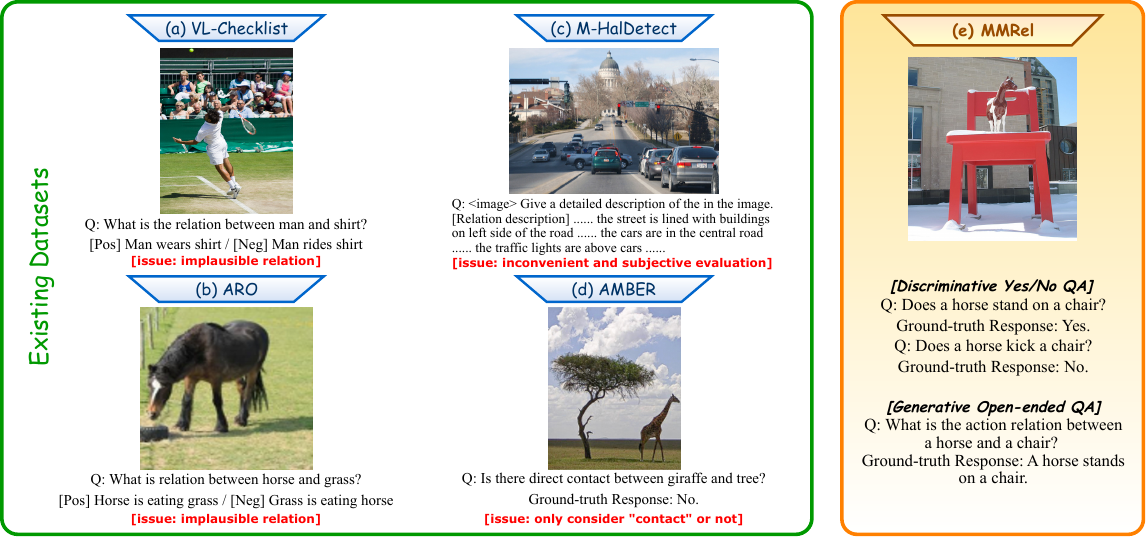}
    \vspace{-4mm}
    \caption{% Limitations of existing benchmarks. (a)-(b) negative choices are implausible and can be easily ruled out; (c) evaluation metrics are complex and subjective; (d) merely considering one type of relation ``contact'', which is insufficient and ambiguous. 
    Limitations of existing benchmarks: Implausible negative choices which can be easily ruled out, as illustrated in (a) and (b); Complex and subjective evaluation metrics as in (c); Incomplete and ambiguous relation annotation of ``contact'' only as in (d). In contrast, MMRel mitigates these limitations by: \textit{\textbf{(i)}} providing a comprehensive taxonomy of inter-object relations, \textit{\textbf{(ii)}} carefully designing plausible negative action relations, and \textit{\textbf{(iii)}} adopting both \textbf{discriminative} \textit{Yes/No} and \textbf{generative} \textit{open-ended} evaluations.}
    % \vspace{+2mm}
\label{fig:prior_dataset}
\end{figure*}

Specifically, VL-Checklist~\cite{zhao2022vl} and ARO~\cite{yuksekgonul2022and} serve as representative benchmarks for evaluating how well MLLMs understand fine-grained visual details, such as inter-object relations. These two benchmarks inherit images and labels from VG~\cite{krishna2017visual} and adopt a binary \textit{Positive/Negative} evaluation framework, as illustrated in Fig.~\ref{fig:prior_dataset}(a)-(b). However, both benchmarks have significant limitations, as their negative choices are made randomly without human verification. Specifically, they generate negative choices by: \textit{(i)} randomly altering and grouping the objects and relations (\textit{e.g.}, man riding shirt); \textit{(ii)} simply reversing the order of subjects and objects (\textit{e.g.}, grass is eating the horse), making the generated negative options semantically implausible or even impossible. With such designs, MLLMs can easily deduce the correct answers even without seeing images, as the powerful LLM enables MLLMs to effectively dismiss those negative choices that starkly deviate from common sense.

Moreover, although recent MLLMs~\cite{liu2023visual,liu2023improved, ye2023mplugowl,li2023otter,dai2023instructblip, gong2023multimodalgpt,bai2023qwen} are versatile, their proficiency in object detection is limited due to pre-training on coarse granularity image-text pairs~\cite{sharma2018conceptual,changpinyo2021conceptual}. Therefore, the overly dense objects, occasionally incorrect annotations, and unclear taxonomies from existing benchmarks (\textit{e.g.}, VG~\cite{krishna2017visual}) are unsuitable for evaluating MLLMs. Consequently, re-labeling existing benchmarks to develop suitable benchmarks is crucial for evaluating relation understanding capabilities of MLLMs. This has triggered recent initiatives such as M-HalDetect~\cite{gunjal2024detecting} and AMBER~\cite{wang2023llm}. Specifically, M-HalDetect~\cite{gunjal2024detecting} incorporates manually annotated relations within lengthy descriptive sentences, with evaluations by human assessors (refer to Fig.~\ref{fig:prior_dataset}(c)). However, this metric does not straightforwardly reflect the capabilities in relation understanding and introduces potential risks of subjective comparisons. On the other hand, AMBER~\cite{wang2023llm} simplifies the classification of relationships to \textit{contact or not} and employs a \textit{Yes/No} evaluation framework (refer to Fig.~\ref{fig:prior_dataset}(d)). While this approach is straightforward, it leads to an imprecise evaluation due to the restricted definition of relations.

\noindent The limitations of existing benchmarks are:\\
\noindent$\bullet$ \textit{\textbf{First}}, they are small-scale and lack diversity.\\
\noindent$\bullet$ \textit{\textbf{Second}}, their annotations are of low quality, with negative choices for action relations often being meaningless.\\
\noindent$\bullet$ \textit{\textbf{Third}}, they lack clear definitions and taxonomies of inter-object relations. Discussions of more benchmarks~\cite{cheng2024spatialrgpt,rasheed2024glamm,liu2023vsr,kamath2023s,shao2024visual} are in Sec.~\ref{ssec:benchmark_supp} of the appendix.

\section{MMRel Benchmark}
\label{sec:mmrel}

As analyzed in Sec.~\ref{sec:review}, existing benchmarks for relation understanding are generally constrained by their small-scale and homogeneous nature. This motivates us to develop a new benchmark, Multi-Modal Relation Understanding (MMRel), characterized by large-scale, diverse, and high-quality data on inter-object relations. In this section, we first present the statistics and question formats of the MMRel (Sec.~\ref{ssec:statistics}). Subsequently, we describe the image sources and relation taxonomies (Sec.~\ref{ssec:data}). Next, we introduce a Semi-automatic Data Collection pipeline, designed to collect large-scale, diverse, and high-quality data (Sec.~\ref{ssec:semidc}). Finally, we discuss some potential concerns about the constructed MMRel (Sec.~\ref{ssec:discussion_mmrel}).

\subsection{Statistics and Question Format of MMRel}
\label{ssec:statistics}

Fig.~\ref{fig:statistics} shows the statistics of MMRel. Specifically, MMRel comprises around 22{,}500 question-answer pairs (15{,}000 \textit{Yes/No}, and 7{,}500 \textit{open-ended}) across 7 subsets, spanning 3 domains and 394 kinds of relations from three categories. Concretely, approximately 68\% of the data consists of real images, while the remaining 32\% comprises synthetic images (\textit{i.e.}, SDXL~\cite{podell2023sdxl} and Dall-E~\cite{betker2023improving} generated images). In terms of relation categories, MMRel includes a larger scale of spatial and action relations compared to comparative relations. Thanks to the open-vocabulary annotation capability of GPT-4V~\cite{achiam2023gpt}, MMRel achieves a diverse range of objects and action relations, as shown in Fig.~\ref{fig:statistics}. Inspired by \cite{bai2024coig}, we recognize that unusual data are crucial for evaluating and enhancing MLLM capabilities, and we create an adversarial relation subset comprising 710 challenging question-answer pairs. These pairs are carefully selected from the full MMRel dataset to cover images across all domains and include every relation category. This subset rigorously assesses the relation understanding capabilities of MLLMs, the statistics of which are described in the following section.

For question format, MMRel incorporates both \textbf{discriminative} \textit{Yes/No} questions and \textbf{generative} \textit{open-ended} questions. Moreover, it is noteworthy that MMRel questions strictly describe the objects and their relations without including any attributes. This ensures the evaluation accurately reflect the relation understanding capabilities. For example, if we introduce attributes into the question \textit{``Is blue bird right to yellow bird?''}, and the MLLMs answers differ from the ground-truth, we cannot ascertain whether the discrepancy stems from a misunderstanding of color (\textit{i.e.}, \textit{blue} and \textit{yellow}) or inter-object relation (\textit{i.e.}, \textit{right}). Even though MMRel questions are simple and straightforward, they still exhibit sufficient complexity and challenges for MLLMs, as discussed in Sec.~\ref{ssec:discussion_mmrel}.

\noindent In summary, MMRel is featured with four distinctive characteristics as illustrated in Fig.~\ref{fig:prior_dataset}(e):\\
\noindent$\bullet$ \textit{\textbf{First}}, it comprises large-scale and diverse relation data that are collected from multiple sources.\\
\noindent$\bullet$ \textit{\textbf{Second}}, it comprises high-quality relation annotations and negative-choice data for action relations.\\
\noindent$\bullet$ \textit{\textbf{Third}}, it adopts an effective and efficient evaluation framework~\cite{li2023evaluating} that employs both \textbf{discriminative} \textit{Yes/No} and \textbf{generative} \textit{open-ended} questions.\\
\noindent$\bullet$ \textit{\textbf{Fourth}}, it includes an adversarial subset to evaluate the MLLMs' ability to understand challenging relations.

\begin{figure*}[t]
\centering
    \vspace{-2mm}
    \includegraphics[width=\linewidth]{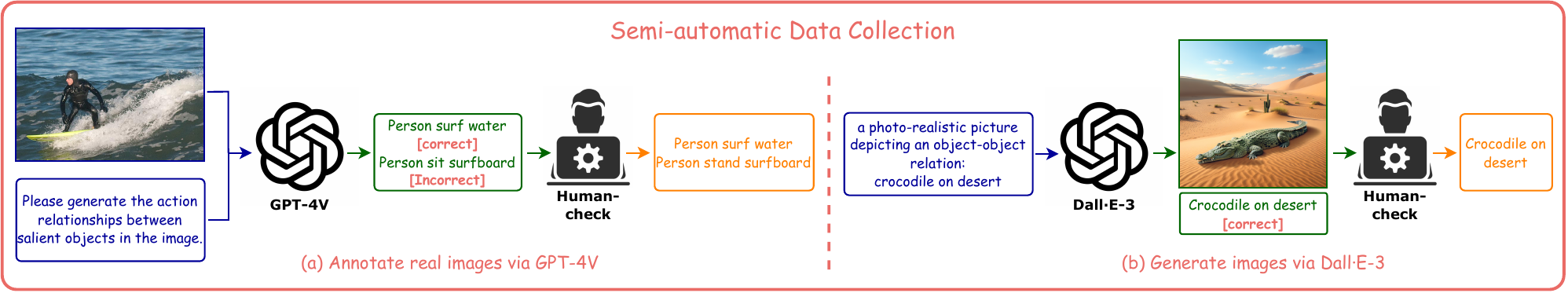}
    \vspace{-4mm}
    \caption{%We propose Semi-automatic Data Collection (SemiDC) pipeline to collect data of MMRel benchmark. (a) Re-labeling images from the Visual Genome~\cite{krishna2017visual} using GPT-4V~\cite{achiam2023gpt}, with human verification. (b) Generating synthetic images via Dall-E~\cite{betker2023improving}, followed by human verification to ensure accuracy.
    We propose a Semi-automatic Data Collection (SemiDC) pipeline that constructs MMRel with two distinctive approaches: (a) re-label images of the Visual Genome~\cite{krishna2017visual} with GPT-4V~\cite{achiam2023gpt}, and (b) generate synthetic images with Dall-E~\cite{betker2023improving}. Both approaches are followed by human verification to ensure the data quality.}
    \vspace{-2mm}
\label{fig:semidc}
\end{figure*}

\subsection{Data Source and Relation Taxonomy}
\label{ssec:data}

Following prior works~\cite{zhao2022vl,yuksekgonul2022and}, we first obtain the real-image subset from Visual Genome (VG)~\cite{krishna2017visual} and then introduce synthetic images into MMRel to create out-of-distribution subsets with respect to the training data. We first select synthetic images from the SPEC~\cite{peng2023synthesize}, which is generated from SDXL~\cite{podell2023sdxl}. Specifically, we use the \textit{relative spatial} subset for spatial relations and the \textit{relative size} subset for comparative relations. The selected images are characterized by two or three objects against a clean background, providing an accurate evaluation of relation understanding capabilities. Moreover, SPEC contains a small portion of counterintuitive comparative relations that contribute to the adversarial subset. We also employ Dall-E~\cite{betker2023improving} to generate synthetic images of different styles to further diversify MMRel. Leveraging the text-to-image generation, we specifically design prompts depicting some counterintuitive spatial and action relations. Details regarding annotation and image generation across all three and all relation categories domains are outlined in Sec.~\ref{ssec:semidc}.

Moreover, we employ distinct annotation strategies for three relation taxonomies. \textit{\textbf{(i)}} For spatial relations, we limit the categories to \textit{left}, \textit{right}, \textit{on/over}, and \textit{under}, as other descriptors such as \textit{around} or \textit{next to} do not accurately capture relative inter-object positions. \textit{\textbf{(ii)}} For action relations, our annotations are based on the capabilities of Multi-Modal Large Language Models (MLLMs). \textit{\textbf{(iii)}} Comparative relations are manually annotated, as MLLMs exhibit limited reliability in this category. Our provided taxonomies of relations are more suitable in the MLLM era compared with prior benchmarks, which is discussed in Sec.~\ref{ssec:relation_supp}.

\subsection{Comparison with Relation in Prior Benchmarks}\label{ssec:relation_supp}

The taxonomy of relations in SGG~\cite{zellers2018neural} is inherited from VG~\cite{krishna2017visual}. In Sec.~\ref{sec:review}, we outline the limitations of VG~\cite{krishna2017visual}, particularly emphasizing its overly-dense annotations and frequent inaccuracies. We argue that these factors render VG suboptimal for use with MLLMs. The differences between the relation taxonomy used in VG and our contributed MMRel are as follows: \textit{\textbf{(i)}} The relations in VG are classified into four categories: \textit{(a)} Geometric: such as behind and under; \textit{(b)} Possessive: such as has, part of, wearing, etc; \textit{(c)} Semantic: such as eating and using; and \textit{(d)} Misc: for, from, and made of. In contrast, relations in MMRel fall into three categories: \textit{(a)} Spatial, \textit{(b)} Action, and \textit{(c)} Comparative. \textit{\textbf{(ii)}} Our taxonomy is similar to that of VG, as we both incorporate spatial (geometric in VG) and action relations (possessive and semantic in VG). Comparative relations in MMRel are absent in VG. \textit{\textbf{(iii)}} For spatial relations, to ensure precision and clarity, we adopt \textit{left, right, on/over, and under}, rather than the more ambiguous expressions, such as \textit{behind}, \textit{next to}, and \textit{in front of} used in VG. \textit{\textbf{(iv)}} Some possessive relations, specifically \textit{has} and \textit{part of}, can cause confusion and lead to inaccuracies for MLLMs due to their imprecise definitions. \textit{\textbf{(v)}} We exclude miscellaneous relations because: \textit{(a)} \textit{made of} should be treated as an attribute rather than a relation, as in VL-Checklist~\cite{zhao2022vl} and PACO~\cite{bianchi2024devil} (\textit{e.g., a cup is made of glass}); \textit{(b)} other relations, such as \textit{for} and \textit{from}, are prone to ambiguity due to their broad semantic interpretations.

\subsection{Semi-Automatic Data Collection Pipeline}
\label{ssec:semidc}

This section presents the Semi-Automatic Data Collection pipeline (SemiDC). As illustrated in Fig.~\ref{fig:semidc}, SemiDC consists of two branches, including one for re-labeling existing images and the other for generating synthetic images.

\begin{figure*}[h]
\centering
   \includegraphics[width=0.96\linewidth]{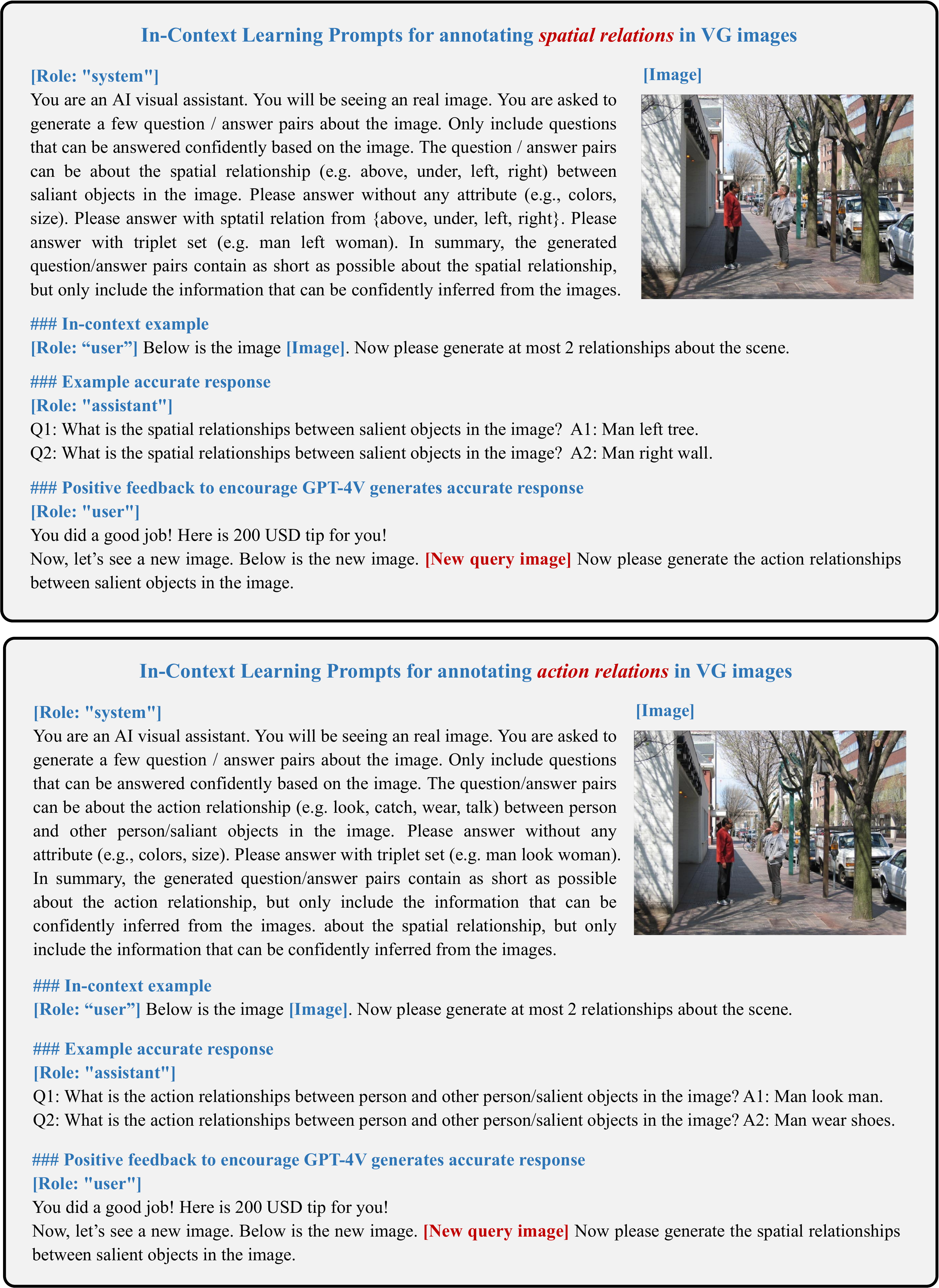}
   \vspace{-1mm}
   \caption{In-context learning prompts for labeling images from Visual Genome~\cite{krishna2017visual}.}
   % \vspace{-2mm}
\label{fig:gpt_prompt}
\end{figure*}

\noindent\textbf{SemiDC for Annotation} Since most existing image annotations are incompatible with MLLMs~\cite{zhao2022vl,yuksekgonul2022and}, we employ GPT-4V~\cite{achiam2023gpt} to generate relation annotations for Visual Genome (VG)~\cite{krishna2017visual}. The process consists of three stages: \textit{\textbf{(i)}} Pre-processing: We filter out complex images with over 10 tiny objects (each $<$1/20 of the total image area) that GPT-4V~\cite{achiam2023gpt} cannot process well; \textit{\textbf{(ii)}} Re-labeling via GPT-4V~\cite{achiam2023gpt}: With in-context learning~\cite{li2023otter,zhao2023mmicl}, we provide reference annotations for each query image with the prompt \textit{``Please generate the action relationships between salient objects in this image.''} The specific prompts used are illustrated in Fig.~\ref{fig:gpt_prompt}. \textit{\textbf{(iii)}} Human verification: We manually review and correct GPT-4V annotations to ensure accuracy. For comparative relations, we construct this subset by manually re-labeling VG images.

Since MLLMs tend to favor positive responses~\cite{li2023evaluating,liu2023mitigating}, we balance \textit{``Yes''} and \textit{``No''} answers to mitigate bias in binary questions. For spatial and comparative relations, negatives are derived by reversing object order (\textit{e.g.}, \textit{``dog is left to cat''} \textit{v.s.} \textit{``cat is left to dog''}). But for action relations, randomly generating negatives is implausible. To address this issue, we use GPT-3.5~\cite{blog2022introducing} to generate plausible negative relations that are aligned with common sense, followed by manual verification (\textit{e.g.}, a negative counterpart for \textit{``man eats cake''} could be \textit{``man bakes cake''}). This improves both training and evaluation over prior work~\cite{zhao2022vl, yuksekgonul2022and}. Additionally, we use LLM for evaluating MLLMs via open-ended generative questions (\textit{e.g.}, \textit{``What is the action relation between a dog and a frisbee?''}). The potential biases arising from GPT-generated annotations are examined in Sec.~\ref{ssec:discussion_mmrel}.

\begin{table*}[!t]

\caption{Evaluation results of LLaVA-1.5~\cite{liu2023improved} and its text-only version on our MMRel benchmark.}
\label{tab:text-only}
\vspace{-1mm}

\centering
\renewcommand\arraystretch{1.25}

\setlength{\tabcolsep}{5.8pt}

\begin{tabular}{l||ccccc|ccccc|ccccc}
\toprule
\multirow{3}{*}{\textbf{Model}} & \multicolumn{15}{c}{\textit{\textbf{Real Domain}}}\\ \cline{2-16}
& \multicolumn{5}{c|}{\textit{\textbf{Spatial}}} & \multicolumn{5}{c|}{\textit{\textbf{Action}}} & \multicolumn{5}{c}{\textit{\textbf{Comparative}}} \\ \cline{2-16}
& \textbf{Acc.} & \textbf{Prec.} & \textbf{Rec.} & \textbf{F1} & \textbf{Yes-Rate (Y.R.)} & \textbf{Acc.} & \textbf{Prec.} & \textbf{Rec.} & \textbf{F1} & \textbf{Y.R.} & \textbf{Acc.} & \textbf{Prec.} & \textbf{Rec.} & \textbf{F1} & \textbf{Y.R.} \\ \hline\hline
\ \ LLaVA-1.5~\cite{liu2023improved} & 51.5 & 51.0 & 74.9 & 60.7 & 73.4 \ & \ 67.7 & 63.0 & 88.4 & 73.5 & 71.3 \ & 61.7 & \ 61.6 & 62.3 & 61.9 & 50.6 \\
\ \ Text-only & 48.2 & 49.4 & 42.8 & 46.7 & 41.6 & 57.0 & 55.9 & 66.6 & 60.8 & 59.6 & 51.0 & 51.6 & 57.0 & 54.2 & 56.1 \\
\bottomrule
\end{tabular}

\vspace{1mm}

\setlength{\tabcolsep}{3.4pt}

\begin{tabular}{l||ccccc|ccccc||ccccc|ccccc}
\toprule
\multirow{3}{*}{\textbf{Model}} & \multicolumn{10}{c||}{\textit{\textbf{SDXL Domain}}} & \multicolumn{10}{c}{\textit{\textbf{Dall-E Domain}}}\\ \cline{2-21}
& \multicolumn{5}{c|}{\textbf{\textit{Spatial}}} & \multicolumn{5}{c||}{\textbf{\textit{Comparative}}} & \multicolumn{5}{c|}{\textbf{\textit{Spatial}}} & \multicolumn{5}{c}{\textbf{\textit{Action}}} \\\cline{2-21}
& \textbf{Acc.} & \textbf{Prec.} & \textbf{Rec.} & \textbf{F1} & \textbf{Y.R.} & \textbf{Acc.} & \textbf{Prec.} & \textbf{Rec.} & \textbf{F1} & \textbf{Y.R.} & \textbf{Acc.} & \textbf{Prec.} & \textbf{Rec.} & \textbf{F1} & \textbf{Y.R.} & \textbf{Acc.} & \textbf{Prec.} & \textbf{Rec.} & \textbf{F1} & \textbf{Y.R.} \\ \hline\hline
LLaVA-1.5~\cite{liu2023improved} & 52.5 & 54.5 & 84.6 & 64.0 & 82.1 \ & \ 59.1 & 58.8 & 60.9 & 59.8 & 51.8 \ & \ 53.2 & 51.9 & 88.9 & 65.5 & 85.7 \ & \ 61.2 & 60.8 & 71.4 & 65.7 & 61.0\ \\
Text-only & 51.4 & 52.3 & 31.6 & 39.4 & 30.2 & 49.5 & 49.6 & 53.5 & 51.5 & 54.0 & 49.3 & 48.7 & 28.6 & 36.0 & 29.3 & 45.0 & 44.6 & 24.1 & 31.3 & 28.1\ \\
\bottomrule
\end{tabular}

\vspace{+3mm}

\end{table*}

\begin{table}[h]
\vspace{+1mm}
\caption{Gemini-2.5~\cite{comanici2025gemini} and GPT-4o~\cite{blog2024hello} performance on two subsets annotated by GPT-4V~\cite{achiam2023gpt}.}
\label{tab:gemini}
\vspace{-1mm}

\centering
\renewcommand\arraystretch{1.25}

\setlength{\tabcolsep}{3.8pt}

\begin{tabular}{l||cccc|cccc}
\toprule
\multirow{2}{*}{\textbf{Model}} & \multicolumn{4}{c|}{\textbf{\textit{Real Spatial}}} & \multicolumn{4}{c}{\textbf{\textit{Real Action}}}\\ \cline{2-9}
& \textbf{Acc.} & \textbf{Prec.} & \textbf{Rec.} & \textbf{F1} & \textbf{Acc.} & \textbf{Prec.} & \textbf{Rec.} & \textbf{F1} \\ \hline\hline
GPT-4o~\cite{blog2024hello} & 59.1 &55.1 &98.1 &70.6 &79.7 &71.9 &98.6 &83.2\\
Gemini-2.5~\cite{comanici2025gemini} & 73.0 &68.5 &85.0 &75.9 &88.1 &91.2 &88.3 &89.7\\ 
Human & \multicolumn{4}{c|}{98.8 (Accuracy)} & \multicolumn{4}{c}{97.7 (Accuracy)} \\ \bottomrule
\end{tabular}

\vspace{+1mm}
    
\end{table}

\noindent\textbf{SemiDC for Generation.} Beyond annotation, SemiDC synthesizes diverse inter-object relations via generative models as illustrated in Fig.~\ref{fig:sample}(b). The generation involves three stages: \textit{\textbf{(i)}} Preparing textual prompts: We define inter-object relations through structured text prompts; \textit{\textbf{(ii)}} Image generation via DALL-E~\cite{betker2023improving}: Prompts (\textit{e.g.}, \textit{``a photo-realistic image of a crocodile on a desert''}) guide image generation. We incorporate four image styles—\textit{photo-realistic}, \textit{watercolor}, \textit{abstract}, and \textit{oil painting}—to enhance the data diversity as illustrated in Fig.~\ref{fig:sample}(c); \textit{\textbf{(iii)}} Human verification: If generated images misalign with the input relations, we adjust the annotations. Similarly, negative action relations are generated using GPT-3.5~\cite{blog2022introducing}. We use the following prompts as inputs for DALL-E-3~\cite{betker2023improving} to generate images: \textit{a \textbf{style} photo depicting a strange or unusual object-object relation: \textbf{inter-object relations}}. Here, \textit{\textbf{style}} specifies the image style, selected from \textit{photo-realistic, watercolor, abstract, and oil painting}, and \textit{\textbf{inter-object relations}} describes the specific relation we aim to depict. More Dall-E generated examples are provided in Fig.~\ref{fig:dall-e_sample} of the appendix. For example, the prompt for generating the first image in Fig.~\ref{fig:dall-e_sample} is: \textit{an \textbf{abstract} photo depicting a strange or unusual object-object relation: \textbf{cat on toilet}}.

\subsection{Discussion on Constructed MMRel}\label{ssec:discussion_mmrel}
\noindent\textbf{MMRel Question Difficulty is Suitable for MLLMs.} The adversarial subset of MMRel, which is manually curated, presents significant challenges for MLLMs. However, this does not imply that the remaining subsets are overly simple. To verify that other MMRel subsets also require genuine visual understanding, rather than common knowledge or text-only cues, we evaluate LLaVA-1.5~\cite{liu2023improved} and its text-only variant, as shown in Tab.~\ref{tab:text-only}. The text-only model performs consistently worse than the baseline, indicating that common-sense priors alone are insufficient for MMRel. Moreover, the baseline model exhibits a strong bias toward answering ``\textit{yes}'', suggesting that it frequently relies on guesswork rather than understanding. Although the text-only model and the baseline achieve similar accuracy on spatial sub-tasks, their underlying error patterns differ substantially. For instance, in the detailed Real-Spatial results, the text-only model yields 1{,}389 false negatives and 983 false positives, whereas the baseline yields 566 false negatives and 1{,}731 false positives. This discrepancy highlights the crucial role of visual information in shaping the model’s decisions. The degraded performance of the text-only setup further validates the quality of our carefully constructed negative samples—these negatives cannot be trivially inferred from common knowledge and indeed require accurate visual understanding.

\noindent\textbf{Risk of Annotation Bias.} The annotations in MMRel pose minimal risk of bias from the annotator models, the evidences are as follow. \textit{\textbf{(i)}} Annotations generated by GPT-4V~\cite{achiam2023gpt} are verified by humans, with approximately 40\% initially found to be inaccurate and subsequently corrected manually. Since the annotations reflect human judgment, MMRel carries a low bias risk and provides reliable evaluation of MLLMs’ capabilities. \textit{\textbf{(ii)}} Although GPT-4o~\cite{blog2024hello} may operate under an in-domain setup, it still performs significantly worse than humans (refer to Tab.~\ref{tab:gemini}), highlighting both the challenge of the relation understanding task and the effectiveness of MMRel. \textit{\textbf{(iii)}} Furthermore, we evaluate Gemini-2.5~\cite{comanici2025gemini} on GPT-4V-annotated subsets (VG-spatial and action) in Tab.~\ref{tab:gemini}. It outperforms GPT-4o~\cite{blog2024hello}, further confirming that the risk of annotation bias is negligible.
\section{Experiment}
\label{sec:exp}

In this section, we systematically analyze the effectiveness of Multi-Modal Relation Understanding benchmark (MMRel). We first describe the experiment setups (Sec.~\ref{ssec:setup}). Subsequently, we employ MMRel to evaluate the relation understanding capabilities of representative Multi-Modal Large Language Models (MLLMs)~(Sec.\ref{ssec:evaluation} and Sec.~\ref{ssec:open-ended}). Finally, we demonstrate that fine-tuning with MMRel enhances relation understanding capabilities (Sec.~\ref{ssec:training}).

\subsection{Experiment Setups}
\label{ssec:setup}
\noindent\textbf{Selected MLLMs.} We evaluate MMRel over \textbf{28} representative MLLMs and relevant methods. Their details are introduced as follows.

\noindent\textbf{General-purpose MLLMs.}  We first choose \textbf{17} open-sourced general-pripose MLLMs: including LLaVA-1.5 series~\cite{liu2023improved}, InstructBLIP~\cite{dai2023instructblip}, Qwen-VL series~\cite{bai2023qwen,qwen2025qwen3}, MiniCPM-V~\cite{yao2024minicpm}, Phi-3.5-Vision~\cite{microsoft2024phi}, LLaVA-OneVision series~\cite{li2024llava}, InternVL-3.5 series~\cite{zhu2025internvl3}, and SAIL-VL2 series~\cite{yin2025sail}. Specifically, LLaVA-1.5~\cite{liu2023improved} and  InstructBLIP~\cite{dai2023instructblip} incorporate Vicuna LLM~\cite{chiang2023vicuna}. MiniCPM-V selects LLaMA LLM~\cite{dubey2024llama}. Phi-3.5-V~\cite{microsoft2024phi} adoptes Phi-3.5 LLM~\cite{microsoft2024phi}. Qwen-VL~\cite{bai2023qwen,qwen2025qwen3}, LLaVA-OneVision~\cite{li2024llava}, InternVL-3.5~\cite{zhu2025internvl3} adopt Qwen LLM~\cite{bai2023qwen}. Nevertheless, they are trained with image-text pairs of coarse granularity~\cite{li2019visualbert,li2022blip} and thus prone to generating hallucinations~\cite{rohrbach2018object, wang2023evaluation}.

\newpage
\begin{table*}[!t]
\caption{%Evaluation results of various MLLMs on the MMRel. Results indicate that most general-purpose MLLMs still struggle with inter-object relation understanding. MLLM with multiple vision~\cite{tong2024cambrian} encoders and Reasoning MLLMs~\cite{xiaomi2025mimo} show significantly strong performances.
Evaluations of general-purpose MLLMs over MMRel. Most general-purpose MLLMs struggle while handling inter-object relation understanding. Including multiple vision encoders~\cite{tong2024cambrian} or reasoning capability~\cite{xiaomi2025mimo} enhances the relation understanding significantly.}
\label{tab:evaluation}
\vspace{-1mm}

\centering
\renewcommand\arraystretch{1.25}

\setlength{\tabcolsep}{9.46pt}

\begin{tabular}{l||cccc|cccc|cccc}
\toprule
\multirow{3}{*}{\textbf{Model}} & \multicolumn{12}{c}{\textbf{\textit{Real Domain}}}\\\cline{2-13}
& \multicolumn{4}{c|}{\textbf{\textit{Spatial}}} & \multicolumn{4}{c|}{\textbf{\textit{Action}}} & \multicolumn{4}{c}{\textbf{\textit{Comparative}}} \\\cline{2-13}
& \textbf{Acc.} & \textbf{Prec.} & \textbf{Rec.} & \textbf{F1} & \textbf{Acc.} & \textbf{Prec.} & \textbf{Rec.} & \textbf{F1} & \textbf{Acc.} & \textbf{Prec.} & \textbf{Rec.} & \textbf{F1} \\\hline\hline
InstructBLIP-7B~\cite{dai2023instructblip} & 50.6 & 50.4 & 81.4 & 62.2 & 65.1 & 60.6 & 89.9 & 72.4 & 56.2 & 55.0 & 69.0 & 61.2 \\
LLaVA-1.5-7B~\cite{liu2023improved} & 50.5 & 50.3 & 75.0 & 60.2 & 67.7 & 63.0 & 88.4 & 73.5 & 61.7 & 61.6 & 62.3 & 61.9 \\
LLaVA-1.5-13B~\cite{liu2023improved} & 54.9 & 52.8 & 93.2 & 67.4 & 78.9 & 72.5 & 94.5 & 82.0 & 73.0 & 94.7 & 48.6 & 64.3 \\
Qwen-VL-7B~\cite{bai2023qwen} & 54.1 & 52.7 & 81.3 & 63.9 & 69.9 & 64.0 & 92.9 & 75.8 & 72.2 & 70.0 & 77.8 & 73.7 \\
MiniCPM-V-2.8B~\cite{yao2024minicpm} & 52.9 & 51.9 & 80.6 & 63.1 & 76.5 & 70.1 & 94.1 & 80.3 & 62.5 & 59.5 & 77.8 & 67.5 \\
Phi-3.5-V-4.2B~\cite{microsoft2024phi} & 69.5 & 63.6 & 91.1 & 74.9 & 72.2 & 66.0 & 93.4 & 77.4 & 82.8 & 78.9 & 89.7 & 83.9 \\
LLaVA-OV-0.5B~\cite{li2024llava} & 50.2 & 50.1 & 89.7 & 64.3 & 65.5 & 60.3 & 94.7 & 73.6 & 55.0 & 54.7 & 58.7 & 56.6 \\
LLaVA-OV-7B~\cite{li2024llava} & 59.2 & 55.2 & 97.5 & 70.5 & 79.0 & 71.2 & 98.5 & 82.6 & 89.2 & 85.4 & 94.5 & 89.8 \\
InternVL-3.5-2B~\cite{zhu2025internvl3} & 57.0 & 53.9 & 97.6 & 69.5 & 73.0 & 67.2 & 96.4 & 79.2 & 66.1 & 83.1 & 85.1 & 84.1 \\
InternVL-3.5-8B~\cite{zhu2025internvl3} & 72.0 & 64.5 & 98.5 & 78.0 & 81.1 & 76.5 & 95.5 & 84.9 & 85.4 & 91.0 & 92.7 & 91.9 \\
InternVL-3.5-38B~\cite{zhu2025internvl3} & 71.9 & 65.9 & 98.4 & 78.9 & 78.6 & 76.5 & 98.2 & 86.0 & 85.7 & 94.8 & 93.3 & 94.0 \\
Qwen3-VL-2B~\cite{qwen2025qwen3} & 66.7 & 60.6 & 95.4 & 74.1 & 73.7 & 66.6 & 96.6 & 78.9 & 86.6 & 82.0 & 93.9 & 87.5 \\
Qwen3-VL-8B~\cite{qwen2025qwen3} & 83.3 & 76.3 & 96.6 & 85.3 & 84.6 & 79.2 & 94.5 & 86.2 & 91.2 & 85.2 & 99.7 & 91.9 \\
% Qwen3-VL-32B~\cite{qwen2025qwen3} & 92.8 & 89.0 & 97.5 & 93.1 & 89.2 & 84.5 & 96.4 & 90.1 & 96.2 & 95.8 & 96.7 & 96.2 \\
SAIL-VL2-2B~\cite{yin2025sail} & 54.0 & 52.1 & 98.6 & 68.2 & 82.7 & 75.7 & 97.0 & 85.1 & 86.9 & 83.5 & 92.1 & 87.6 \\
SAIL-VL2-8B~\cite{yin2025sail} & 73.2 & 65.4 & 98.4 & 78.6 & 86.6 & 80.2 & 97.9 & 88.2 & 94.8 & 91.6 & 98.8 & 95.0 \\
InternVL-3.5-A3B~\cite{zhu2025internvl3} & 62.0 & 57.0 & 97.7 & 72.0 & 78.6 & 72.0 & 98.2 & 83.1 & 49.4 & 98.5 & 77.8 & 86.9 \\
Kimi-VL-A3B~\cite{team2025kimi} & 54.5 & 52.6 & 90.9 & 66.7 & 79.1 & 72.1 & 96.3 & 82.4 & 90.3 & 88.0 & 93.3 & 90.6 \\
Cambrian-8B~\cite{tong2024cambrian} & 75.2 & 70.9 & 85.5 & 77.5 & 80.7 & 74.7 & 93.9 & 83.2 & 88.2 & 94.0 & 81.5 & 87.3 \\
% Cambrian-13B~\cite{tong2024cambrian} & 65.0 & 60.3 & 87.5 & 71.4 & 80.0 & 74.0 & 93.6 & 82.6 & 87.4 & 89.9 & 84.2 & 87.0 \\
MiMo-VL-7B~\cite{xiaomi2025mimo} & 84.0 & 86.0 & 89.9 & 87.9 & 88.3 & 88.1 & 92.3 & 90.2 & 93.2 & 95.9 & 92.7 & 94.3 \\
MiMo-VL-7B-RL~\cite{xiaomi2025mimo} & 83.4 & 87.2 & 88.7 & 88.0 & 88.1 & 88.8 & 91.7 & 90.3 & 92.9 & 95.6 & 91.8 & 93.6 \\
Janus-Pro-7B~\cite{chen2025janus} & 69.5 & 69.3 & 70.0 & 69.7 & 75.7 & 70.3 & 90.3 & 79.1 & 69.6 & 81.8 & 50.5 & 62.4 \\
GPT-4o~\cite{blog2024hello} & 59.1 & 55.1 & 98.1 & 70.6 & 79.7 & 71.9 & 98.6 & 83.2 & 94.8 & 93.3 & 96.7 & 94.9 \\ \bottomrule
\end{tabular}

\vspace{1mm}

\setlength{\tabcolsep}{5.16pt}

\begin{tabular}{l||cccc|cccc||cccc|cccc}
\toprule
\multirow{3}{*}{\textbf{Model}} & \multicolumn{8}{c||}{\textbf{\textit{SDXL Domain}}} & \multicolumn{8}{c}{\textbf{\textit{Dall-E Domain}}} \\\cline{2-17}
& \multicolumn{4}{c|}{\textbf{\textit{Spatial}}} & \multicolumn{4}{c||}{\textbf{\textit{Comparative}}} & \multicolumn{4}{c|}{\textbf{\textit{Spatial}}} & \multicolumn{4}{c}{\textbf{\textit{Action}}} \\\cline{2-17}
& \textbf{Acc.} & \textbf{Prec.} & \textbf{Rec.} & \textbf{F1} & \textbf{Acc.} & \textbf{Prec.} & \textbf{Rec.} & \textbf{F1} & \textbf{Acc.} & \textbf{Prec.} & \textbf{Rec.} & \textbf{F1} & \textbf{Acc.} & \textbf{Prec.} & \textbf{Rec.} & \textbf{F1} \\ \hline\hline
InstructBLIP-7B~\cite{dai2023instructblip} & 51.4 & 50.9 & 75.7 & 60.9 & 51.2 & 51.0 & 65.9 & 57.5 & 55.3 & 53.1 & 91.7 & 67.2 & 63.1 & 61.3 & 79.0 & 69.0 \\
LLaVA-1.5-7B~\cite{liu2023improved} & 51.4 & 50.9 & 80.7 & 62.4 & 59.1 & 58.8 & 60.9 & 59.8 & 53.0 & 51.7 & 90.1 & 65.7 & 60.3 & 60.3 & 69.1 & 64.4 \\
LLaVA-1.5-13B~\cite{liu2023improved} & 52.1 & 51.1 & 97.9 & 67.1 & 62.0 & 81.0 & 31.4 & 45.2 & 51.1 & 50.6 & 97.9 & 66.7 & 69.9 & 64.6 & 93.3 & 76.3 \\
Qwen-VL-7B~\cite{bai2023qwen} & 55.8 & 54.6 & 69.5 & 61.1 & 70.2 & 77.6 & 56.6 & 65.5 & 57.4 & 55.0 & 82.2 & 65.9 & 62.5 & 62.9 & 67.8 & 65.3 \\
MiniCPM-V-2.8B~\cite{yao2024minicpm} & 52.4 & 51.6 & 80.1 & 62.7 & 58.5 & 55.8 & 81.9 & 66.3 & 60.4 & 56.0 & 96.9 & 71.0 & 70.9 & 75.7 & 64.9 & 69.9 \\
Phi-3.5-V-4.2B~\cite{microsoft2024phi} & 67.6 & 61.3 & 95.5 & 74.7 & 76.8 & 79.0 & 72.9 & 75.8 & 54.2 & 52.3 & 97.4 & 68.0 & 71.9 & 69.6 & 81.3 & 75.0 \\
LLaVA-OV-0.5B~\cite{li2024llava} & 49.5 & 49.7 & 76.2 & 60.1 & 49.7 & 46.2 & 3.7 & 6.8 & 53.1 & 51.7 & 92.2 & 66.3 & 60.0 & 57.6 & 87.8 & 69.5 \\
LLaVA-OV-7B~\cite{li2024llava} & 60.4 & 55.8 & 99.1 & 71.4 & 84.2 & 81.9 & 87.7 & 84.7 & 51.6 & 50.8 & 99.5 & 67.3 & 75.3 & 69.0 & 95.5 & 80.1 \\
InternVL-3.5-2B~\cite{zhu2025internvl3} & 65.6 & 59.8 & 94.7 & 73.3 & 68.8 & 62.6 & 96.6 & 75.9 & 52.6 & 51.4 & 99.0 & 67.6 & 67.9 & 64.1 & 89.3 & 74.6 \\
InternVL-3.5-8B~\cite{zhu2025internvl3} & 81.6 & 73.6 & 98.7 & 84.3 & 64.9 & 91.9 & 52.3 & 66.7 & 60.3 & 56.8 & 98.5 & 72.1 & 77.1 & 76.0 & 87.8 & 81.5 \\
InternVL-3.5-38B~\cite{zhu2025internvl3} & 80.2 & 75.6 & 91.8 & 82.9 & 59.9 & 72.1 & 70.2 & 71.1 & 61.4 & 58.0 & 98.8 & 73.1 & 63.2 & 85.6 & 70.4 & 77.3 \\
Qwen3-VL-2B~\cite{qwen2025qwen3} & 71.5 & 66.4 & 87.1 & 75.3 & 71.4 & 71.0 & 72.3 & 71.6 & 52.7 & 51.5 & 95.4 & 66.9 & 75.9 & 71.9 & 88.0 & 79.2 \\
Qwen3-VL-8B~\cite{qwen2025qwen3} & 86.4 & 81.6 & 94.0 & 87.4 & 74.5 & 72.0 & 80.0 & 75.8 & 85.0 & 78.0 & 97.7 & 86.7 & 80.4 & 87.3 & 72.9 & 79.4 \\
% Qwen3-VL-32B~\cite{qwen2025qwen3} & 89.8 & 86.5 & 94.3 & 90.2 & 79.1 & 83.2 & 72.9 & 77.7 & 91.9 & 87.9 & 97.2 & 92.3 & 82.6 & 89.3 & 75.6 & 81.9 \\
SAIL-VL2-2B~\cite{yin2025sail} & 53.2 & 51.6 & 100.0 & 68.1 & 66.3 & 60.4 & 94.5 & 73.7 & 50.2 & 50.1 & 99.5 & 66.6 & 79.9 & 76.8 & 87.8 & 81.9 \\
SAIL-VL2-8B~\cite{yin2025sail} & 67.0 & 60.3 & 99.8 & 75.2 & 74.1 & 67.6 & 92.9 & 78.2 & 55.0 & 52.6 & 100.0 & 69.0 & 80.1 & 75.4 & 91.5 & 82.7 \\
InternVL-3.5-A3B~\cite{zhu2025internvl3} & 79.3 & 71.3 & 98.3 & 82.7 & 54.6 & 88.8 & 58.5 & 70.5 & 62.7 & 60.7 & 98.2 & 75.1 & 66.9 & 82.5 & 68.1 & 74.6 \\
Kimi-VL-A3B~\cite{team2025kimi} & 62.8 & 58.7 & 85.8 & 69.7 & 75.2 & 73.0 & 80.0 & 76.4 & 57.4 & 54.1 & 96.7 & 69.4 & 79.8 & 78.8 & 83.6 & 81.1 \\
Cambrian-8B~\cite{tong2024cambrian} & 82.7 & 85.4 & 78.8 & 82.0 & 79.5 & 99.0 & 59.7 & 74.5 & 67.3 & 61.4 & 93.2 & 74.0 & 73.7 & 77.0 & 70.4 & 73.6 \\
% Cambrian-13B~\cite{tong2024cambrian} & 63.0 & 58.7 & 87.4 & 70.3 & 78.9 & 97.0 & 59.7 & 73.9 & 60.4 & 56.0 & 97.2 & 71.1 & 72.4 & 75.0 & 70.4 & 72.7 \\
MiMo-VL-7B~\cite{xiaomi2025mimo} & 88.0 & 86.0 & 92.3 & 89.1 & 66.8 & 72.4 & 58.8 & 64.9 & 83.6 & 83.1 & 94.9 & 88.6 & 82.0 & 89.5 & 76.4 & 82.5 \\
MiMo-VL-7B-RL~\cite{xiaomi2025mimo} & 87.2 & 87.3 & 92.1 & 89.6 & 65.5 & 74.2 & 55.7 & 63.6 & 85.0 & 85.4 & 93.9 & 89.4 & 80.6 & 89.8 & 74.5 & 81.4 \\
Janus-Pro-7B~\cite{chen2025janus} & 74.9 & 75.6 & 73.5 & 74.5 & 74.1 & 72.9 & 76.9 & 74.8 & 63.5 & 59.0 & 88.6 & 70.8 & 77.9 & 78.6 & 78.9 & 78.7 \\
GPT-4o~\cite{blog2024hello} & 79.7 & 71.9 & 98.6 & 83.2 & 71.0 & 74.6 & 68.6 & 71.5 & 75.2 & 67.2 & 98.7 & 79.9 & 81.1 & 78.1 & 88.3 & 82.9 \\ \bottomrule
\end{tabular}

\vspace{+1mm}

\setlength{\tabcolsep}{48.4pt}

\begin{tabular}{l||cccc|cccc|cccc}
\toprule
\multirow{2}{*}{\textbf{Model}} & \multicolumn{4}{c|}{\textit{\textbf{Spatial}}} & \multicolumn{4}{c|}{\textit{\textbf{Action}}} & \multicolumn{4}{c}{\textit{\textbf{Comparative}}} \\ \cline{2-13}
~ & \multicolumn{4}{c|}{\textbf{Acc.}} & \multicolumn{4}{c|}{\textbf{Acc.}} & \multicolumn{4}{c}{\textbf{Acc.}} \\ \hline \hline
Human & \multicolumn{4}{c|}{98.8} & \multicolumn{4}{c|}{97.7} & \multicolumn{4}{c}{98.1} \\ \bottomrule
\end{tabular}

\end{table*}

\clearpage
\newpage

\begin{table*}[ht]

\caption{%Evaluation results of grounding-focused MLLMs on MMRel indicate that these models still struggle with inter-object relation understanding.
Evaluations of grounding-focused MLLMs over MMRel. Grounding-focused MLLMs struggle with inter-object relation understanding.}
\label{tab:evaluation_ground}
\vspace{-1mm}

\centering
\renewcommand\arraystretch{1.25}

\setlength{\tabcolsep}{10.2pt}

\begin{tabular}{l||cccc|cccc|cccc}
\toprule
\multirow{3}{*}{\textbf{Model}} & \multicolumn{12}{c}{\textbf{\textit{Real Domain}}}\\\cline{2-13}
& \multicolumn{4}{c|}{\textbf{\textit{Spatial}}} & \multicolumn{4}{c|}{\textbf{\textit{Action}}} & \multicolumn{4}{c}{\textbf{\textit{Comparative}}} \\\cline{2-13}
& \textbf{Acc.} & \textbf{Prec.} & \textbf{Rec.} & \textbf{F1} & \textbf{Acc.} & \textbf{Prec.} & \textbf{Rec.} & \textbf{F1} & \textbf{Acc.} & \textbf{Prec.} & \textbf{Rec.} & \textbf{F1} \\\hline\hline
Shikra~\cite{chen2023shikra} & 50.9 & 50.5 & 82.0 & 62.5 & 70.8 & 65.0 & 92.3 & 76.3 & 62.2 & 59.0 & 79.9 & 67.9 \\
Ferret~\cite{you2023ferret} & 46.9 & 48.0 & 73.1 & 57.9 & 56.1 & 54.6 & 81.2 & 65.3 & 53.6 & 54.3 & 45.9 & 49.8 \\
SpatialRGPT~\cite{cheng2024spatialrgpt} & 57.0 & 57.9 & 80.6 & 67.4 & 58.3 & 55.7 & 87.8 & 68.2 & 69.6 & 66.8 & 77.8 & 71.9 \\ \bottomrule
\end{tabular}

\vspace{1mm}

\setlength{\tabcolsep}{5.88pt}

\begin{tabular}{l||cccc|cccc||cccc|cccc}
\toprule
\multirow{3}{*}{\textbf{Model}} & \multicolumn{8}{c||}{\textbf{\textit{SDXL Domain}}} & \multicolumn{8}{c}{\textbf{\textit{Dall-E Domain}}} \\\cline{2-17}
& \multicolumn{4}{c|}{\textbf{\textit{Spatial}}} & \multicolumn{4}{c||}{\textbf{\textit{Comparative}}} & \multicolumn{4}{c|}{\textbf{\textit{Spatial}}} & \multicolumn{4}{c}{\textbf{\textit{Action}}} \\\cline{2-17}
& \textbf{Acc.} & \textbf{Prec.} & \textbf{Rec.} & \textbf{F1} & \textbf{Acc.} & \textbf{Prec.} & \textbf{Rec.} & \textbf{F1} & \textbf{Acc.} & \textbf{Prec.} & \textbf{Rec.} & \textbf{F1} & \textbf{Acc.} & \textbf{Prec.} & \textbf{Rec.} & \textbf{F1} \\ \hline\hline
Shikra~\cite{chen2023shikra} & 51.4 & 50.8 & 87.4 & 64.3 & 64.3 & 61.1 & 78.8 & 68.8 & 49.4 & 49.7 & 91.1 & 64.3 & 66.2 & 63.7 & 81.3 & 71.4 \\
Ferret~\cite{you2023ferret} & 41.6 & 43.8 & 59.2 & 50.3 & 48.0 & 48.2 & 52.6 & 50.3 & 49.8 & 49.9 & 81.8 & 62.0 & 57.1 & 57.3 & 68.5 & 62.4 \\
SpatialRGPT~\cite{cheng2024spatialrgpt} & 58.2 & 54.5 & 97.7 & 70.0 & 62.6 & 59.1 & 81.9 & 68.6 & 58.2 & 54.5 & 97.7 & 70.0 & 59.4 & 58.5 & 75.6 & 66.0 \\ \bottomrule
\end{tabular}

\end{table*}
\begin{table*}[ht]

\vspace{+2mm}

\caption{%Evaluation results of various hallucination mitigation methods on the MMRel. Results indicate that these methods still struggle with inter-object relation understanding.
Evaluations of hallucination mitigation methods over MMRel. Hallucination mitigation methods still struggle with inter-object relation understanding}
\label{tab:evaluation_hall}
\vspace{-1mm}

\centering
\renewcommand\arraystretch{1.25}

\setlength{\tabcolsep}{10.88pt}

\begin{tabular}{l||cccc|cccc|cccc}
\toprule
\multirow{3}{*}{\textbf{Model}} & \multicolumn{12}{c}{\textbf{\textit{Real Domain}}}\\ \cline{2-13}
& \multicolumn{4}{c|}{\textbf{\textit{Spatial}}} & \multicolumn{4}{c|}{\textbf{\textit{Action}}} & \multicolumn{4}{c}{\textbf{\textit{Comparative}}} \\ \cline{2-13}
& \textbf{Acc.} & \textbf{Prec.} & \textbf{Rec.} & \textbf{F1} & \textbf{Acc.} & \textbf{Prec.} & \textbf{Rec.} & \textbf{F1} & \textbf{Acc.} & \textbf{Prec.} & \textbf{Rec.} & \textbf{F1} \\ \hline\hline
VCD~\cite{leng2023mitigating} & 50.8 & 50.4 & 92.0 & 65.1 & 65.3 & 59.9 & 96.6 & 73.9 & 62.2 & 60.6 & 69.3 & 64.7 \\
DOLA~\cite{chuang2023dola} & 51.5 & 51.0 & 78.5 & 61.8 & 69.1 & 63.5 & 92.1 & 75.2 & 60.6 & 60.5 & 61.4 & 60.9 \\
OPERA~\cite{huang2023opera} & 51.2 & 50.7 & 90.2 & 64.9 & 70.4 & 63.7 & 96.7 & 76.8 & 74.3 & 75.6 & 71.7 & 73.6 \\
\bottomrule
\end{tabular}

\vspace{1mm}

\setlength{\tabcolsep}{6.4pt}

\begin{tabular}{l||cccc|cccc||cccc|cccc}
\toprule
\multirow{3}{*}{\textbf{Model}} & \multicolumn{8}{c||}{\textbf{\textit{SDXL Domain}}} & \multicolumn{8}{c}{\textbf{\textit{Dall-E Domain}}} \\\cline{2-17}
& \multicolumn{4}{c|}{\textbf{\textit{Spatial}}} & \multicolumn{4}{c||}{\textbf{\textit{Comparative}}} & \multicolumn{4}{c|}{\textbf{\textit{Spatial}}} & \multicolumn{4}{c}{\textbf{\textit{Action}}} \\\cline{2-17}
& \textbf{Acc.} & \textbf{Prec.} & \textbf{Rec.} & \textbf{F1} & \textbf{Acc.} & \textbf{Prec.} & \textbf{Rec.} & \textbf{F1} & \textbf{Acc.} & \textbf{Prec.} & \textbf{Rec.} & \textbf{F1} & \textbf{Acc.} & \textbf{Prec.} & \textbf{Rec.} & \textbf{F1} \\ \hline\hline
VCD~\cite{leng2023mitigating} & 50.6 & 50.3 & 96.7 & 66.2 & 64.3 & 63.2 & 68.6 & 65.8 & 51.6 & 50.9 & 97.9 & 66.9 & 61.6 & 58.7 & 88.8 & 70.6 \\
DOLA~\cite{chuang2023dola} & 52.1 & 51.2 & 89.7 & 65.2 & 58.5 & 57.5 & 64.9 & 61.0 & 63.0 & 61.4 & 77.6 & 68.5 & 53.1 & 51.7 & 94.1 & 66.7 \\
OPERA~\cite{huang2023opera} & 50.9 & 50.5 & 97.3 & 66.5 & 67.9 & 65.3 & 76.0 & 70.3 & 69.2 & 64.7 & 89.9 & 74.2 & 50.4 & 50.2 & 97.9 & 66.4 \\ \bottomrule
\end{tabular}

\end{table*}

\noindent\textbf{Mixture-of-Experts-based MLLMs.} We additionally include two MoE-based MLLMs for evaluation: InstructVL-3.5-A3B~\cite{zhu2025internvl3} and Kimi-VL-A3B~\cite{team2025kimi}. We hypothesize that certain activated experts specialized in fine-grained visual understanding may provide benefits for relation understanding.

\noindent\textbf{MLLM with Multiple Vision Encoders.} As relation understanding is a visually-centric task, we also evaluate Cambrian-8B~\cite{tong2024cambrian}. It incorporates multiple vision encoders~\cite{oquab2023dinov2,radford2021learning,cherti2023reproducible} to enhance performance on visual-centric tasks, and we expect it to be beneficial. Cambrian-8B~\cite{tong2024cambrian} adopts LLaMA-Instruct-8B~\cite{dubey2024llama} as its LLM.

\noindent\textbf{Reasoning MLLMs.} To further increase the comprehensiveness of our evaluation and analysis, we also assess the relation-understanding capabilities of MiMo MLLMs~\cite{xiaomi2025mimo}, which exhibit enhanced thinking and reasoning abilities in responding a long chain of verifications before answering ``\textit{yes}'' or ``\textit{no}''.

\noindent\textbf{Unified Understanding and Generation MLLM.} Motivated by recent advances in unified multi-modal understanding and generation MLLMs, we further evaluate Janus-Pro-7B~\cite{chen2025janus} on MMRel. Given that the model encodes relation descriptions in prompts during image generation, we expect it to exhibit strong performance on MMRel as well.

\noindent\textbf{Proprietary MLLM.} Furthermore, we include the latest proprietary model, GPT-4o~\cite{blog2024hello}, which has demonstrated remarkable capabilities across various vision-language tasks.

\noindent\textbf{Grounding-focused MLLMs.} Moreover, we evaluate three grounding-focused MLLMs on MMRel: Shikra~\cite{chen2023shikra}, Ferret~\cite{you2023ferret}, and SpatialRGPT~\cite{cheng2024spatialrgpt}. Since these models exhibit stronger object-grounding capabilities, we hypothesize that they may offer advantages in relation understanding. Shikra~\cite{chen2023shikra} and Ferret~\cite{you2023ferret} use Vicuna-7B~\cite{chiang2023vicuna} as their language decoder, while SpatialRGPT adopts the LLaMA-7B backbone~\cite{dubey2024llama}.

\noindent\textbf{Hallucination Mitigation Methods.} We also include three methods designed for hallucination mitigation: DOLA~\cite{chuang2023dola}, OPERA~\cite{huang2023opera}, and VCD~\cite{leng2023mitigating}. Since these methods aim to reduce hallucinations, we hypothesize that they may be better suited for understanding inter-object relations. The evaluated versions of DOLA~\cite{chuang2023dola}, OPERA~\cite{huang2023opera}, and VCD~\cite{leng2023mitigating} are all built upon LLaVA-1.5-7B~\cite{liu2023improved}.

\noindent\textbf{Evaluation Merics.} For \textbf{discriminative} \textit{Yes/No} evaluation, we adopt the same metrics, accuracy, precision, recall, and F1-Score, as in POPE~\cite{li2023evaluating}. For \textbf{generative} \textit{open-ended} questions, our evaluations are assisted by LLM and follow the LLaVA-Bench~\cite{liu2023visual} framework.

\subsection{MLLM Evaluations with MMRel Yes/No Questions}
\label{ssec:evaluation}
We employ all of the 15{,}000 \textbf{discriminative} \textit{Yes/No} questions in MMRel to evaluate how MLLMs perform while understanding inter-object relations. As Tab.~\ref{tab:evaluation}, Tab.~\ref{tab:evaluation_ground}, and Tab.~\ref{tab:evaluation_hall} show, most MLLMs face various problems while handling relation understanding tasks. Several points can be drawn from the experimental results\footnote{We would like to clarify that this paper presents general conclusions based on common trends across most cases, while acknowledging that some MLLMs may exhibit slightly different results.}.

\noindent\textbf{Discussion of Three Relation Types.} Regarding the three types of inter-object relations in MMRel, we observe that most general-purpose MLLMs perform clearly better on action and comparative relations for real images. The improved performance can be largely attributed to the coarsely aligned MLLM training data, which predominantly capture action and comparative descriptions rather than spatial terms, as the latter often involve vague expressions such as \textit{near}, \textit{next to}, \textit{in front of}, and so on.

\noindent\textbf{Discussion of Three Image Domains.} From the perspective of synthetic \textit{v.s.} real data, we observe that most general-purpose MLLMs generally perform better on real images compared to synthetic images generated by SDXL~\cite{podell2023sdxl} (for comparative relations) and DALL-E~\cite{betker2023improving} (for action relations). We attribute this to the huge-scale training data. The performance gap is largely attributed to the distribution discrepancy between the synthetic images and the real images used in MLLM training. For spatial relations, MLLMs are more likely to produce ``\textit{random}'' responses based on language priors (vague expressions), and synthetic images with clean backgrounds are easier than real ones, beneficial for similar or higher results. Furthermore, Dall-E images exhibit art styles and are distinct from the training data. Thus, MLLMs rely more on images rather than language priors, leading to improved performance.

\noindent\textbf{Discussion of Model Size.} As shown in Tab.~\ref{tab:evaluation}, we observe that the scaling law generally holds in MMRel: larger models within the same series consistently outperform their smaller counterparts. This trend is evident across multiple series, including LLaVA-1.5~\cite{liu2023improved}, LLaVA-OneVision~\cite{li2024llava}, Intern-VL-3.5~\cite{zhu2025internvl3}, Qwen3-VL~\cite{qwen2025qwen3}, and SAIL-VL2~\cite{yin2025sail}. In addition, smaller MLLMs, such as Phi-3.5-V-4.2B~\cite{microsoft2024phi}, LLaVA-OneVision-0.5B~\cite{li2024llava}, and SAIL-VL2-2B~\cite{yin2025sail}, exhibit constrained performance.

\noindent\textbf{Discussion of Recent MLLMs.} As shown in Tab.~\ref{tab:evaluation}, we observe that early genernal-prupose MLLMs, \textit{i.e.}, LLaVA-1.5~\cite{liu2023improved}, InstructBLIP~\cite{dai2023instructblip}, and Qwen-VL~\cite{bai2023qwen}, struggle with relation understanding. In contrast, their subsequent versions, such as LLaVA-OneVision~\cite{li2024llava} and Qwen3-VL~\cite{qwen2025qwen3}, achieve substantially better results. Moreover, other recent general-purpose MLLMs, such as InternVL-3.5~\cite{zhu2025internvl3}, also demonstrate strong capabilities: their performance even surpasses that of GPT-4o~\cite{blog2024hello}. This trend suggests that large-scale training data, along with well-designed architectures and training strategies, can significantly enhance MLLMs’ relation understanding abilities.

\noindent\textbf{Evaluation of MoE-based MLLMs.} We observe that MoE-based MLLMs do not exhibit clear performance gains over other models of similar scale (see Tab.~\ref{tab:evaluation}). This highlights the difficulty of relation understanding and suggests that merely activating visual-centric experts is insufficient to address this challenge.

\begin{figure}[t]
\centering
% \vspace{-2mm}
   \includegraphics[width=\linewidth]{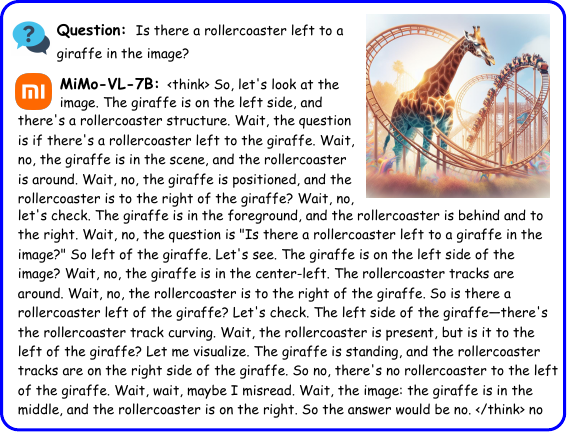}
   \vspace{-4mm}
   \caption{Reasoning MLLMs generate a long chain of verifications before producing the final answer.}
\label{fig:reason}
\end{figure}

\begin{table*}[t]

\vspace{+2mm}

\caption{LLM-assisted open-ended evaluation of some MLLMs on our MMRel benchmark. The full mark is 10.}
\label{tab:llm-eval}
\vspace{-1mm}

\centering
\renewcommand\arraystretch{1.25}

\setlength{\tabcolsep}{14.4pt}

\begin{tabular}{l||ccc||cc||cc}
\toprule
\multirow{2}{*}{\textbf{Model}} & \multicolumn{3}{c||}{\textbf{\textit{Real Domain}}} & \multicolumn{2}{c||}{\textbf{\textit{SDXL Domain}}} & \multicolumn{2}{c}{\textbf{\textit{Dall-E Domain}}} \\ \cline{2-8}
& \textbf{\textit{Spatial}} & \textbf{\textit{Action}} & \textbf{\textit{Comparative}} & \textbf{\textit{Spatial}} & \textbf{\textit{Comparative}} & \textbf{\textit{Spatial}} & \textbf{\textit{Action}} \\ \hline\hline
InstructBLIP~\cite{dai2023instructblip} & 1.3 & 5.3 & 5.0 & 1.2 & 5.5 & 2.0 & 4.8 \\
LLaVA-1.5~\cite{liu2023improved} & 1.3 & 5.3 & 5.1 & 1.2 & 5.0 & 2.0 & 4.8 \\
Qwen-VL~\cite{bai2023qwen} & 1.2 & 5.3 & 5.3 & 1.1 & 5.3 & 2.0 & 4.8 \\
\bottomrule
\end{tabular}

\vspace{-2mm}

\end{table*}

\noindent\textbf{Evaluation of MLLM with Multiple Vision Encoders.} In Tab.~\ref{tab:evaluation}, we observe that Cambrian-8B~\cite{tong2024cambrian}, which leverages multiple vision encoders, surpasses most other MLLMs by large margins. This supports our hypothesis that relation understanding is a vision-centric task and that employing multiple encoders is beneficial.

\noindent\textbf{Evaluation of Reasoning MLLMs.} To our surprise, the MiMo-VL series~\cite{xiaomi2025mimo} exhibits exponentially stronger performance in relation understanding, even surpassing much larger MLLMs and GPT-4o~\cite{blog2024hello} (see Tab.~\ref{tab:evaluation}). This underscores that reasoning, a key focus of recent research, is crucial for fine-grained visual understanding tasks. However, as illustrated in Fig.~\ref{fig:reason}, reasoning MLLMs generate a long chain of verifications before producing the final answer, which incurs significantly higher inference cost and may reduce their effectiveness in real-time applications.

\noindent\textbf{Evaluation of Unified MLLM.} In Tab.~\ref{tab:evaluation}, we observe that Janus-Pro-7B~\cite{chen2025janus} achieves strong performance as expected. This suggests that recent progress in unified understanding and generation MLLMs is highly promising.

\noindent\textbf{Discussion of Proprietary MLLMs.} With respect to open-sourced \textit{v.s.} proprietary MLLMs, we observe that GPT-4o~\cite{blog2024hello} significantly outperforms most contemporary open-source MLLMs: LLaVA-1.5 series~\cite{liu2023improved}, InstructBLIP~\cite{dai2023instructblip}, and Qwen-VL series~\cite{bai2023qwen}. This highlights the critical role of large-scale training data and model size in relation understanding, underscoring the great potential for further advancements in open-sourced MLLMs. Notably, we review the annotations generated by GPT-4V~\cite{achiam2023gpt} and find that approximately 40\% annotations contain errors. This highlights that the GPT-4o's better performance is largely due to GPT-4o itself instead of the annotation pipeline. We demonstrate in Sec.~\ref{ssec:discussion_mmrel} that the high performance of GPT-4o~\cite{blog2024hello} does not result from GPT-4V-biased annotations. The evaluations show that GPT-4o does not produce satisfactory performance, though it performs better than many other MLLMs. We also evaluate Chain-of-Thought and In-context techniques with GPT-4o as discussed in Sec.~\ref{ssec:tech_supp} of the appendix.

\begin{figure*}[t]
\centering
\vspace{+4mm}
   \includegraphics[width=\linewidth]{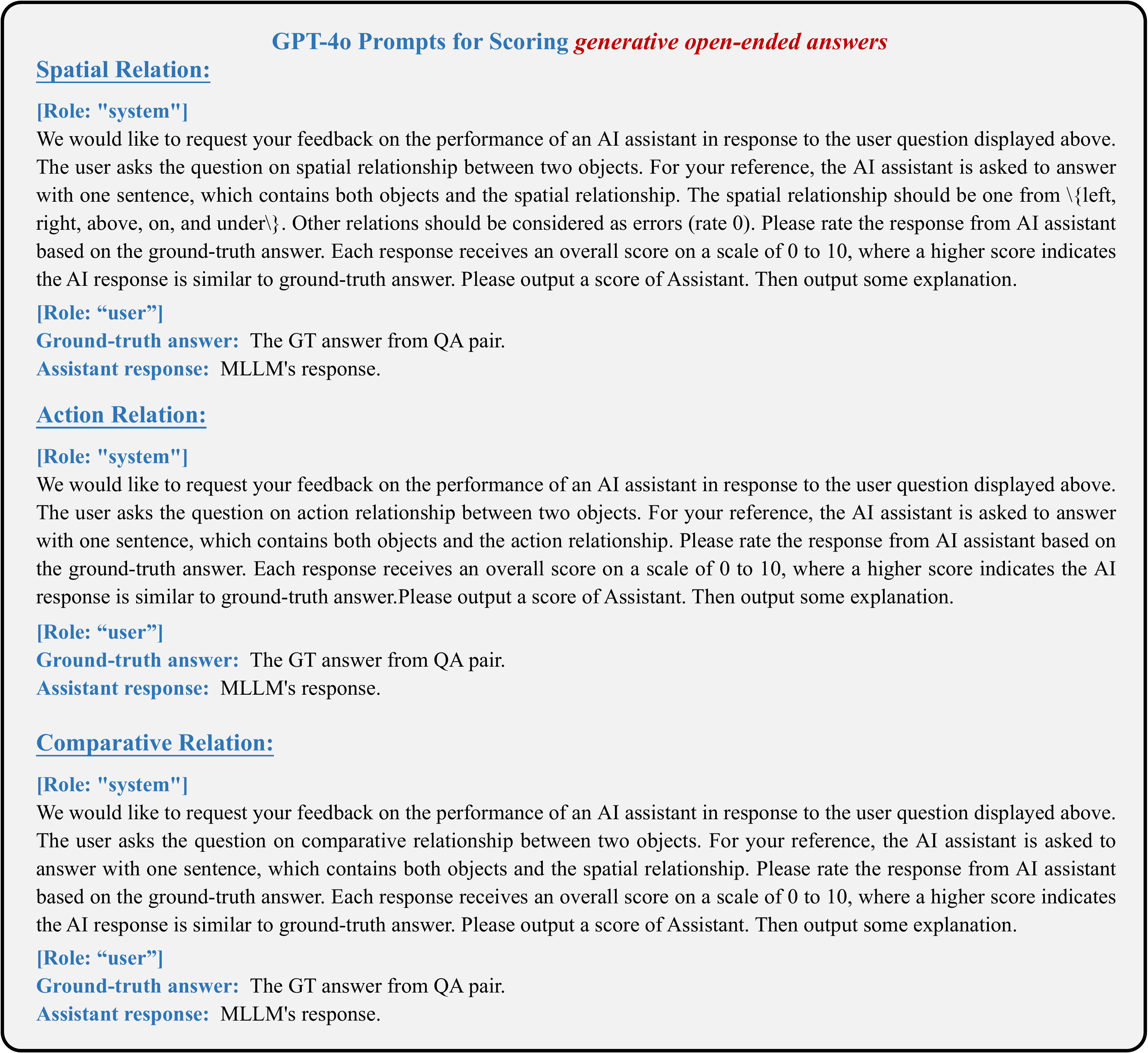}
   \vspace{-4mm}
   \caption{The promptes used for scoring \textbf{generative} \textit{open-ended} answers.}
\label{fig:gpt4_prompt}
\end{figure*}

\noindent\textbf{Human Performance.} We conduct a user study as an upper bound, as reported in Tab.~\ref{tab:evaluation}. Specifically, we randomly sample 900 questions (300 spatial, 300 action, and 300 comparative) and recruit three volunteers to provide answers. Human performance is substantially higher than that of all evaluated MLLMs. From this observation, we draw two conclusions: \textit{\textbf{(i)}} human verification remains essential for reliable data annotation; and \textit{\textbf{(ii)}} there is still considerable room for improvement in relation understanding.

\noindent\textbf{Evaluation of Grounding-focused MLLMs.} As discussed in Sec.~\ref{sec:mmrel}, MMRel strictly consists of objects and their inter-object relations. Since most general-purpose MLLMs struggle with relation understanding, it is important to determine whether the core challenge lies in object grounding or relation understanding. Therefore, we select three grounding-focused MLLMs~\cite{chen2023shikra,you2023ferret,cheng2024spatialrgpt} for evaluation. As shown in Tab.~\ref{tab:evaluation_ground}, these MLLMs still exhibit unsatisfactory performance on MMRel, without showing clear improvements over concurrent general-purpose MLLMs~\cite{liu2023improved,dai2023instructblip,bai2023qwen}. We derive two conclusions from this observation: \textit{\textbf{(i)}} relation understanding cannot be easily improved simply by incorporating training data with bounding boxes; and \textit{\textbf{(ii)}} relation understanding remains a valuable and necessary direction for advancing future MLLMs.

\noindent\textbf{Evaluation of Hallucination Mitigation Methods.} Regarding recent methods on hallucination mitigation~\cite{chuang2023dola,huang2023opera,leng2023mitigating}, we observe that all three evaluated methods remain problematic when applied to MMRel (refer to Tab.~\ref{tab:evaluation_hall}). Although these approaches are effective for object detection, they still struggle with relation understanding, indicating that the primary challenges in MMRel arise from inter-object relationships rather than object hallucination. Moreover, we observe that some versatile MLLMs~\cite{liu2023improved,bai2023qwen,dai2023instructblip} and hallucination mitigation methods achieve low precision but high recall. This suggests that MLLMs' tendency to answer ``\textit{yes}'' over ``\textit{no}'' still persists, futher validating the superiority of our use of balanced ``\textit{yes}'' and ``\textit{no}'' choices.

\begin{table*}[ht]

\caption{Statistics of training, test and adversarial (a challenging subset for testing) splits when MMRel is used for fine-tuning. \textbf{Upper:} number of images in each split. \textbf{Lower:} number of distinct objects and relations in each split.} \label{tab:split}    
\vspace{-1mm}

\centering
\renewcommand\arraystretch{1.25}

    \setlength{\tabcolsep}{3.3pt}

    \begin{tabular}{l||cccc||ccc||ccc||c}    
    \toprule
     \textbf{Domain} &\multicolumn{4}{c||}{\textit{\textbf{Real Domain}}} &\multicolumn{3}{c||}{\textit{\textbf{SDXL Domain}}} &\multicolumn{3}{c||}{\textit{\textbf{Dall-E Domain}}} &\multirow{2}*{\textbf{Total}}\\ \cline{1-11}
     \textbf{Taxonomy} &\textbf{\textit{Spatial}} &\textbf{\textit{Action}} &\textbf{\textit{Comparative}} &\textbf{Total} &\textbf{\textit{Spatial}} &\textbf{\textit{Comparative}} &\textbf{Total} &\textbf{\textit{Spatial}} &\textbf{\textit{Action}} &\textbf{Total} &\\ \hline\hline
     \textbf{\#QA in training split} &$\sim$3.70K &$\sim$3.61K &$\sim$0.59K &$\sim$7.91K &$\sim$1.43K &$\sim$0.45K &$\sim$1.88K &$\sim$0.81K &$\sim$0.79K &$\sim$1.60K &$\sim$11.39K\\ \hline
     \textbf{\#QA in test split} &$\sim$1.52K &$\sim$1.01K &$\sim$0.07K &$\sim$2.60K &$\sim$0.32K &$\sim$0.20K &$\sim$0.52K &$\sim$0.40K &$\sim$0.39K &$\sim$0.79K &$\sim$3.91K\\ \hline
     \textbf{\#QA in adversarial split} &
     \multicolumn{4}{c||}{$\sim$0.67K}& \multicolumn{3}{c||}{$\sim$0.18K}& \multicolumn{3}{c||}{$\sim$0.10K}& $\sim$0.71K\\ \bottomrule 
    \end{tabular}

    \vspace{1mm}

    \setlength{\tabcolsep}{3.0pt}

    \begin{tabular}{l||cc|cc|cc||cc|cc||cc|cc}  
    \toprule
     \textbf{Domain} &\multicolumn{6}{c||}{\textit{\textbf{Real Domain}}} &\multicolumn{4}{c||}{\textit{\textbf{SDXL Domain}}} &\multicolumn{4}{c}{\textit{\textbf{Dall-E Domain}}}\\ \cline{1-15}
     \textbf{Taxonomy} &\multicolumn{2}{c|}{\textbf{\textit{Spatial}}} &\multicolumn{2}{c|}{\textbf{\textit{Action}}} &\multicolumn{2}{c||}{\textbf{\textit{Comparative}}} &\multicolumn{2}{c|}{\textbf{\textit{Spatial}}} &\multicolumn{2}{c||}{\textbf{\textit{Comparative}}} &\multicolumn{2}{c|}{\textbf{\textit{Spatial}}} &\multicolumn{2}{c}{\textbf{\textit{Action}}}\\ \hline
     \textbf{Property} &Object &Relation &Object &Relation &Object &Relation &Object &Relation &Object &Relation &Object &Relation &Object &Relation\\  \hline\hline
     \textbf{Training split} &475 &5 &111 &304 &149 &9 &75 &4 &40 &2 &164 &5 &151 &141\\ \hline
     \textbf{Test split} &226 &5 &111 &304 &39 &9 &52 &4 &28 &2 &133 &5 &136 &141\\ \bottomrule
    \end{tabular}

\end{table*}

\subsection{MLLM Evaluations with Open-ended Questions}\label{ssec:open-ended}

\textbf{Motivation.} Although \textbf{discriminative} \textit{Yes/No} evaluation is widely adopted for its simiplicity, evaluating \textbf{generative} \textit{open-ended} answers is increasingly favored, as it better reflects natural conversational settings. Therefore, we further introduce an LLM-assisted \textit{open-ended} evaluation for relation understanding, making MMRel more comprehensive.

\noindent\textbf{Question-Answer Pair.} Different from the simple questions ``\textit{Is there a cat left to a dog?}'' for \textit{Yes/No} evaluation, \textit{open-ended} questions are more challenging with the format ``\textit{What is the spatial relation between a cat and a dog?}''. We provide more instructions to guide MLLMs to respond with more accurate answers. Specific instructions for three relation categories are as follows:\\
\textit{\textbf{Spatial: Q:} What is the spatial relation between a cat and a dog in the image? Please answer with a sentence containing both objects and relation, such as a dog left to a cat. The spatial relation should be one of \{left, right, on, and under\}. \textbf{A:} a cat is left to a dog.}\\
\textit{\textbf{Action: Q:} What is the action relation between a man and a bay in the image? Please answer with a sentence containing both objects and relation, such as a dog chase a ball. \textbf{A:} a man carries a bag.}\\
\textit{\textbf{Comparative: Q:} What is the comparative relation between a cup and a vase in the image? Please answer with a sentence containing both objects and relation, such as a dog is bigger than a cat. The spatial relation should be one of \{bigger, smaller, taller, longer, and shorter\}. \textbf{A:} a cup is taller than a vase.}

\noindent\textbf{LLM-assisted Evaluation.} We adopt GPT-4o~\cite{blog2024hello} to score the responses from MLLMs (range \textit{0} to \textit{10}), which is similar to LLaVA-Bench~\cite{liu2023visual}.

\noindent{\textbf{GPT-4o Prompts Used for Scoring.}} The specific prompts are illustrated in Fig.~\ref{fig:gpt4_prompt}.

\noindent\textbf{Experiments and Analysis.} As discussed in Sec.~\ref{ssec:evaluation}, grounding-based MLLMs~\cite{chen2023shikra,you2023ferret,cheng2024spatialrgpt} and hallucination mitigation methods~\cite{leng2023mitigating,huang2023opera,chuang2023dola} still struggle with relation understanding. Therefore, we further evaluate three representative general-purpose MLLMs~\cite{liu2023improved,dai2023instructblip,bai2023qwen} using \textbf{generative} \textit{open-ended} questions.

Based on the results presented in Tab.~\ref{tab:llm-eval}, we draw the following conclusions: \textit{\textbf{(i)}} Spatial relations are notably more challenging than action and comparative relations. This may be because training data typically describe coarse spatial interactions (\textit{e.g.}, \textit{near}, \textit{behind}, \textit{next to}), whereas MMRel requires more precise descriptions (\textit{left}, \textit{right}, \textit{on}, and \textit{under}). \textbf{\textit{(ii)}} Action relations in the DALL-E Domain are more difficult than those in the Real Domain, likely due to the domain gap between synthetic images and the training distribution. \textbf{\textit{(iii)}} MLLMs generally exhibit weak inter-object relation understanding. These observations are consistent with the findings from the \textit{Yes/No} evaluations in Sec.~\ref{ssec:evaluation}, demonstrating the effectiveness and fairness of the proposed LLM-assisted open-ended evaluation.

However, the performance differences among the evaluated MLLMs remain small, making it difficult to clearly distinguish their relative capabilities. We attribute this limitation to two factors: \textit{\textbf{(i)}} The evaluator LLM, GPT-4o~\cite{blog2024hello}, is not an oracle for relation understanding, as discussed in Sec.~\ref{sec:exp}; and \textit{\textbf{(ii)}} LLM-assisted evaluations and open-ended responses inherently introduce subjectivity and provide weaker supervision compared with the clarity of \textit{Yes/No} questions. Nevertheless, our LLM-assisted evaluation offers a valuable direction and reference for assessing \textit{open-ended} responses, going beyond the constraints of purely \textit{Yes/No} evaluations. We believe that \textit{open-ended} evaluations represent an important trend, although their effectiveness is currently limited by the capabilities of present-day LLMs.

\begin{table*}[t]
\caption{Fine-tuning LLaVA-1.5~\cite{liu2023improved} with MMRel improves the relation understanding capability significantly.}
\label{tab:ft}
\vspace{-1mm}

\centering
\renewcommand\arraystretch{1.25}

\setlength{\tabcolsep}{3.28pt}

\begin{tabular}{l||ccccc|ccccc|ccccc||ccccc}
\toprule
\multirow{3}{*}{\textbf{Model}} & \multicolumn{15}{c||}{\textbf{\textit{Real Domain}}} & \multicolumn{5}{c}{\textbf{\textit{Adversarial Subset}}} \\\cline{2-21}
& \multicolumn{5}{c|}{\textbf{\textit{Spatial}}} & \multicolumn{5}{c|}{\textbf{\textit{Action}}} & \multicolumn{5}{c||}{\textbf{\textit{Comparative}}} & \multicolumn{5}{c}{\textbf{\textit{Mix}}} \\\cline{2-21}
& \textbf{Acc.} & \textbf{Prec.} & \textbf{Rec.} & \textbf{F1} & \textbf{\#FP$_{\downarrow}$} & \textbf{Acc.} & \textbf{Prec.} & \textbf{Rec.} & \textbf{F1} & \textbf{\#FP$_{\downarrow}$} & \textbf{Acc.} & \textbf{Prec.} & \textbf{Rec.} & \textbf{F1} & \textbf{\#FP$_{\downarrow}$} & \textbf{Acc.} & \textbf{Prec.} & \textbf{Rec.} & \textbf{F1} & \textbf{\#FP$_{\downarrow}$} \\\hline\hline
LLaVA-1.5~\cite{liu2023improved} & 51.3 & 50.9 & 74.8 & 60.6 & 419.0 & 65.1 & 61.7 & \textbf{85.0} & 71.5 & 274.0 & 64.0 & \textbf{83.3} & 66.2 & 73.8 & \textbf{9.0} 
& 50.4 & 47.1 & 70.5 & 56.5 & 255.0 \\
Finetune & \textbf{79.7} & \textbf{82.5} & \textbf{75.3} & \textbf{78.7} & \textbf{93.0} & \textbf{81.2} & \textbf{81.2} & 82.5 & \textbf{81.8} & \textbf{99.0} & \textbf{76.4} & 79.8 & \textbf{92.7} & \textbf{85.7} & 16.0 & \textbf{67.5} & \textbf{60.8} & \textbf{81.4} & \textbf{69.6} & \textbf{169.0} \\
\bottomrule
\end{tabular}

\vspace{1mm}

\setlength{\tabcolsep}{3.3pt}

\begin{tabular}{l||ccccc|ccccc||ccccc|ccccc}
\toprule
\multirow{3}{*}{\textbf{Model}} & \multicolumn{10}{c||}{\textbf{\textit{SDXL Domain}}} & \multicolumn{10}{c}{\textbf{\textit{Dall-E Domain}}} \\\cline{2-21}
& \multicolumn{5}{c|}{\textbf{\textit{Spatial}}} & \multicolumn{5}{c||}{\textbf{\textit{Comparative}}} & \multicolumn{5}{c|}{\textbf{\textit{Spatial}}} & \multicolumn{5}{c}{\textbf{\textit{Action}}} \\\cline{2-21}
& \textbf{Acc.} & \textbf{Prec.} & \textbf{Rec.} & \textbf{F1} & \textbf{\#FP$_{\downarrow}$} & \textbf{Acc.} & \textbf{Prec.} & \textbf{Rec.} & \textbf{F1} & \textbf{\#FP$_{\downarrow}$} & \textbf{Acc.} & \textbf{Prec.} & \textbf{Rec.} & \textbf{F1} & \textbf{\#FP$_{\downarrow}$} & \textbf{Acc.} & \textbf{Prec.} & \textbf{Rec.} & \textbf{F1} & \textbf{\#FP$_{\downarrow}$} \\ \hline\hline
LLaVA-1.5~\cite{liu2023improved} & 50.6 & 50.4 & 80.6 & 62.0 & 127.0 & 61.2 & \textbf{82.9} & 63.0 & 71.6 & \textbf{26.0} & 52.7 & 51.6 & 89.6 & 65.5 & 170.0 & 57.7 & 57.9 & 66.8 & 62.1 & 98 \\
Finetune & \textbf{95.6} & \textbf{96.2} & \textbf{95.0} & \textbf{95.6} & \textbf{6.0} & \textbf{77.1} & 77.4 & \textbf{99.5} & \textbf{87.1} & 58.0 & \textbf{94.1} & \textbf{95.0} & \textbf{93.1} & \textbf{94.0} & \textbf{10.0} & \textbf{71.8} & \textbf{67.0} & \textbf{89.6} & \textbf{76.7} & \textbf{89} \\
\bottomrule
\end{tabular}

\end{table*}

\begin{table*}[ht]

\vspace{+2mm}

\caption{Fine-tuning LLaVA-1.5~\cite{liu2023improved} with MMRel improves performances clearly on MME~\cite{fu2023mme}.}
\label{tab:mme}
\vspace{-1mm}

\centering

\renewcommand\arraystretch{1.25} 

\setlength{\tabcolsep}{6pt}

\begin{tabular}{l||c|ccccccccc|c}
    \toprule
    \textbf{Method} &\textbf{\textit{Position}}	&\textbf{Existence}	&\textbf{Count}	&\textbf{Color}	&\textbf{Posters}	&\textbf{Celebrity}	&\textbf{Scene}	&\textbf{Landmark}	&\textbf{Artwork}	&\textbf{OCR}	&\textbf{Total Score} \\\hline\hline
    LLaVA-1.5~\cite{liu2023improved} &\textit{114.0} &175.7	&124.7	&151.0	&127.8	&113.6	&148.3	&130.0	&102.2	&92.0	&1279.3 \\
    Finetune &\textbf{\textit{126.7}}	&\textbf{180.0}	&\textbf{128.3}	&\textbf{153.3}	&\textbf{129.6}	&\textbf{122.1}	&\textbf{149.3}	&\textbf{131.9}	&\textbf{117.0}	&\textbf{117.5}	&\textbf{1355.7} \\ \bottomrule
\end{tabular}

\vspace{-2mm}

\end{table*}

\begin{figure}[t]
\centering
\vspace{-2mm}
   \includegraphics[width=\linewidth]{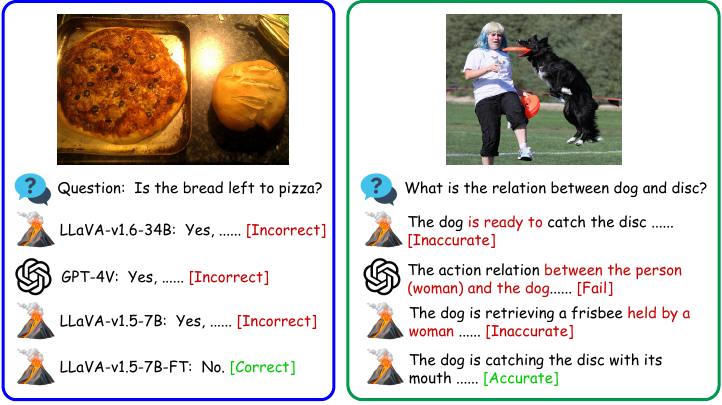}
   \vspace{-4mm}
   \caption{Fine-tuning with MMRel helps to generate more precise and accurate descriptions.}
\label{fig:ft}
\end{figure}

\subsection{Fine-tuning with MMRel}
\label{ssec:training}

\noindent\textbf{Train-test Splits.} The large-scale and diverse data in MMRel enable effective fine-tuning of MLLMs to enhance their relation understanding abilities. To this end, we partition the 15{,}000 \textit{Yes/No} QA paris an 11{,}000 training set and a 4{,}000 test set. The statistics of the split data are summarized in Tab.~\ref{tab:split}. In addition, we collect a subset of adversarial relations, as discussed in Sec~\ref{sec:mmrel}. This subset is manually selected from data across different domains and relation categories (refer to Fig.~\ref{fig:statistics}) and is designed to evaluate MLLMs’ relation understanding under challenging scenarios.

\noindent\textbf{Setup.} We adopt LLaVA-1.5~\cite{liu2023improved} as the baseline due to its popularity, outstanding performance, and versatility.

\noindent\textbf{Fine-tuning Data.} During fine-tuning, we combine the MMRel training data with LLaVA instruction data to to preserve generality and mitigate overfitting to MMRel. Specifically, we integrate the MMRel training set with 260K instances selected from the 665K instruction-tuning data used for LLaVA~\cite{liu2023improved}. This strategy helps preserve the general versatility of MLLMs and prevents overfitting to our specialized dataset. The 260K subset consists of question–answer instruction data drawn from multiple sources: 73K from COCO~\cite{lin2014microsoft}, 53K from VG~\cite{krishna2017visual}, 72K from GQA~\cite{hudson2019gqa}, 22K from TextVQA~\cite{singh2019towards}, and 41K text-only samples.

\noindent\textbf{Question-Answer Format.} The original annotations are in the form of triplets, such as \textit{``cat on car''}. We expand these into both discriminative and generative question-answer pairs. The \textit{\textbf{discriminative}} instructions are as follows:\\
\textit{\textbf{Positive: Q:} Is there a cat on a car in the image? Please answer with one word.} \textit{\textbf{A:} Yes.}\\
\textit{\textbf{Negative: Q:} Is there a car on a cat in the image? Please answer with one word.} \textit{\textbf{A:} No.}\\
The \textit{\textbf{generative}} instructions are as follows:\\
\textit{\textbf{Q:} What is the spatial relation between a cat and a car in the image? \textbf{A:} A cat is on a car.}

\begin{figure}[t]
    \centering
    \vspace{-3mm}
    \includegraphics[width=\linewidth]{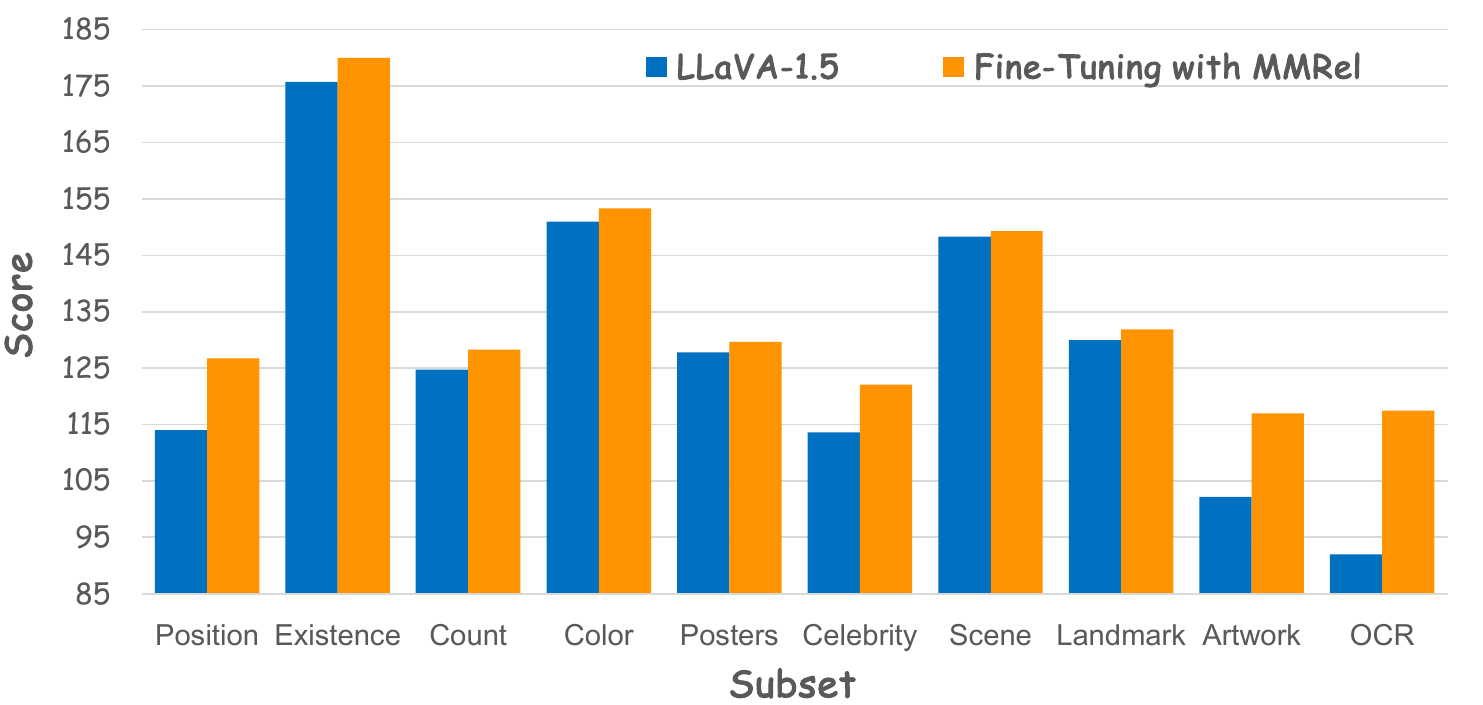}
    \vspace{-4mm}
    \caption{Fine-tuning LLaVA-1.5~\cite{liu2023improved} with MMRel improves performances clearly on MME~\cite{fu2023mme}, validating MMRel benefits other related perceptual tasks.}
    \label{fig:mme}
    \vspace{-2mm}
\end{figure}

\noindent\textbf{Loss.} The loss function used for fine-tuning is the same as LLaVA: $p\left(\mathbf{X}_{\mathrm{a}} \mid \mathbf{X}_{\mathrm{v}}, \mathbf{X}_{\text {q}}\right)=\prod_{i=1}^{L} p_{\boldsymbol{\theta}}\left(x_{i} \mid \mathbf{X}_{\mathrm{v}}, \mathbf{X}_{\text {q},<i}, \mathbf{X}_{\mathrm{a},<i}\right)$, where $\mathbf{X}$ represents the input tokens of different types. $\boldsymbol{\theta}$ is the trainable parameters, and $\mathbf{X}_{,<i}$ represents the instruction and answer tokens in all turns before the current prediction token.

\noindent\textbf{Results.} As Tab.~\ref{tab:ft} shows, fine-tuning with MMRel improves the capabilities of relation understanding significantly and consistently across all domains and relation categories. Moreover, it also improves the performance on the adversarial subset as well. Specifically, the narrow definition of hallucinations refers to scenarios where MLLMs respond with ``\textit{yes}'' while the ground truth is ``\textit{no}'', which is identified as ``False Positive (FP).'' The reductions in ``FP'' across most subsets in Tab.~\ref{tab:ft} indicate that fine-tuning with MMRel suppresses hallucinations effectively. Moreover, fine-tuning with MMRel also mitigates the MLLMs’ bias toward answering ``\textit{yes}'' rather than ``\textit{no}''~\cite{li2023evaluating,liu2023mitigating}. For instance, the ``\textit{yes}'' ratio (in the Real Domain) for spatial and action relations decreases to 45.68\% and 52.28\%, respectively, from the original 73.53\% and 70.93\%. We also fine-tune the baseline with different subsets and analyze the results in Sec.~\ref{ssec:ft_supp} of the appendix. Fine-tuning on a stronger MLLM baseline is also in Sec.~\ref{ssec:ft_supp} of the appendix.

\noindent\textbf{Qualitative examples.} We visualize how the fine-tuned model response to examples in Fig.~\ref{fig:ft}. The results illustrate that fine-tuning with MMRel helps to generate more precise and accurate descriptions.

\noindent\textbf{Discussion.} To verify that fine-tuning MMRel benefits other perception tasks, we conduct experiments on an extensive benchmark MME~\cite{fu2023mme}. MME~\cite{fu2023mme} employs a \textit{Yes/No} evaluation framework. Adhering to the original settings, we calculate the final score using the sum of \textit{accuracy} and \textit{accuracy+}, where \textit{accuracy} is determined for each individual question and \textit{accuracy+} is assessed for each image, requiring correct answers to both associated questions. \textit{Accuracy+} serves as a stricter metric, offering a more comprehensive reflection of the MLLMs' capabilities. As Fig.~\ref{fig:mme} Tab.~\ref{tab:mme} shows, the fine-tuning with MMRel improves most MME tasks as compared with the LLaVA-1.5~\cite{liu2023improved}. This verifies the applicability of the proposed MMRel when it serves as fine-tuning data. Please refer to the appendix for more details regarding the evaluation metrics and experimental results.

\section{Key Take-aways}\label{sec:takeaway}
We summarize the key take-aways from this paper: \\
\noindent \textbf{From the Perspective of Dataset Construction:} \\
\noindent$\bullet$ \textbf{\textit{First}}, designing unambiguous questions and plausible negative choices is crucial for evaluating the relation understanding ability of MLLMs. \\
\noindent$\bullet$ \textbf{\textit{Second}}, synthetic images are crucial as they are controllable with fewer distractions and provide diverse domains for evaluating the robustness of MLLMs. \\ 
\noindent$\bullet$ \textbf{\textit{Third}}, even powerful GPT-4V and GPT-4o struggle with understanding inter-object relations, highlighting the necessity of human verification and incorporating human evaluations as an upper bound. \\
\noindent \textbf{From the Perspective of Experiments:} \\
\noindent$\bullet$ \textbf{\textit{First}}, domain gaps and relation types have a large impact on MLLM relation understanding, offering domain-level and category-level insights for future research. \\
\noindent$\bullet$ \textbf{\textit{Second}}, recent MLLMs demonstrate stronger performance, suggesting that large-scale training data and well-designed strategies are crucial for enhancing their capabilities. \\
\noindent$\bullet$ \textbf{\textit{Third}}, MLLMs with multiple vision encoders, reasoning-focused MLLMs, and unified MLLMs demonstrate significantly superior capabilities, indicating that recent explorations in MLLMs are promising. \\
\noindent$\bullet$ \textbf{\textit{Fourth}}, stronger object dection and grounding capabilities do not necessarily improve relation understanding capabilities of MLLMs. \\
\noindent$\bullet$ \textbf{\textit{Fifth}}, fine-tuned MLLMs exhibit strong generalization ability across different relation types.\\
\noindent \textbf{From the Perspective of Future Work:} \\
\noindent$\bullet$ \textbf{\textit{First}}, a carefully designed evaluation metric is essential for assessing generative open-ended questions.\\
\noindent$\bullet$ \textbf{\textit{Second}}, contributing a benchmark with larger scale and more diverse inter-object relation data is a promising research direction.

\section{Conclusion}

We notice that Multi-Modal Large Language Models (MLLMs) encounter difficulties in understanding inter-object relations, a research area that remains under-explored. To address this, we contribute the Multi-Modal Relation Understanding benchmark (MMRel) with a clear taxonomy of inter-object relations and the definition of relation hallucinations. MMRel is specifically designed to probe MLLMs' capabilities on relation understanding and includes a challenging adversarial subset featuring highly unusual relations. Thanks to its large-scale, high-quality, and diverse data, MMRel serves as a versatile tool for both evaluating and fine-tuning MLLMs, enhancing their capabilities in relation understanding and other perceptual vision-language tasks. Extensive experiments and in-depth analysis valaidate the effectiveness of MMRel.
\vspace{30mm}

{
\bibliographystyle{IEEEtran}
\bibliography{references}
}
\vspace{20mm}

% {
% \bibliographystyle{revision_ref}
% \bibliography{references}
% }

% \input{section/X_biography}

% \input{section/X_appendix}

\end{document}